\documentclass[lettersize,journal]{IEEEtran}
\usepackage{csquotes}
\MakeOuterQuote{"}
\usepackage{amsmath,amsfonts}
\usepackage{algorithm}
\usepackage[noend]{algpseudocode}
\usepackage{array}
\usepackage[caption=false,font=normalsize,labelfont=sf,textfont=sf]{subfig}
\usepackage{textcomp}
\usepackage{stfloats}
\usepackage{url}
\usepackage{verbatim}
\usepackage{graphicx}
\usepackage{cite}
\usepackage{microtype}
\usepackage{graphicx}
\usepackage{booktabs} 
\usepackage{hyperref}
\usepackage{amsmath}
\usepackage{amssymb}
\usepackage{mathtools}
\usepackage{amsthm}

\newcommand{\lp}{\left(}
\newcommand{\rp}{\right)}

\newcommand{\ls}{\left[}
\newcommand{\rs}{\right]}

\usepackage[capitalize,noabbrev]{cleveref}
\usepackage[T1]{fontenc}

\theoremstyle{plain}

\theoremstyle{definition}

\theoremstyle{remark}

\begin{document}

\title{Boosting Reinforcement Learning and Planning with Demonstrations: A Survey}

\author{Tongzhou Mu, Hao Su
\thanks{Manuscript received March DD, 2023; revised MM, DD, 2023. \textit{(Corresponding author: Tongzhou Mu.)}}
\thanks{Tongzhou Mu and Hao Su are with the Department of Computer Science and Engineering, University of California, San Diego, CA, USA (e-mail: t3mu@eng.ucsd.edu; haosu@eng.ucsd.edu).}
}

\maketitle

\begin{abstract}

Although reinforcement learning has seen tremendous success recently, this kind of trial-and-error learning can be impractical or inefficient in complex environments.
The use of demonstrations, on the other hand, enables agents to benefit from expert knowledge rather than having to discover the best action to take through exploration. 
In this survey, we discuss the advantages of using demonstrations in sequential decision making, various ways to apply demonstrations in learning-based decision making paradigms (for example, reinforcement learning and planning in the learned models), and how to collect the demonstrations in various scenarios. 
Additionally, we exemplify a practical pipeline for generating and utilizing demonstrations in the recently proposed ManiSkill robot learning benchmark.

\end{abstract}



\section{Introduction}

\IEEEPARstart{R}{einforcement} learning (RL) and planning are two popular methods to solving sequential decision making problems. 
Both approaches, as well as their combination, made remarkable progress in solving difficult tasks (e.g., Go \cite{alphago, alphago_zero}, video games \cite{dqn, openai_five}, robot control \cite{hwangbo2019learning}) when given the advantages of deep learning. 
RL algorithms allow agents to learn through trial and error, adjusting their behavior based on the consequences of their actions. Planning algorithms, on the other hand, involve the use of a model of the environment to make decisions based on the predicted outcomes of different actions.
While traditional planning techniques demand a model as an input, many recent works \cite{pets, planet, muzero, efficient_zero, td_mpc} learn a model by interacting with environments and then plan in the learned model, which is very reminiscent of the RL framework. As a result, nowadays the distinction between RL and planning is hazy.

Typically, deep RL algorithms require tremendous training samples, resulting in very high sample complexity, which refers to the number of samples required for learning an approximately optimal policy. Particularly, unlike the supervised learning paradigm that learns from historically labeled data, typical RL algorithms require the interaction data by running the latest policy in the environment. Once the policy updates, the underlying data distribution (formally the occupancy measure \cite{syed2008apprenticeship}) changes, and the data has to be collected again by running the policy.
This also applies to the planning methods that learn the model interactively.
As such, RL algorithms with high sample complexity are hard to be directly applied in real-world tasks, where trial-and-errors can be highly costly.

However, many domains, such as autonomous driving \cite{sun2020scalability} and robotics \cite{dasari2019robonet} contain a wealth of pre-collected demonstrations for various tasks, which can be leveraged to boost reinforcement learning and planning agents. 
One key benefit of using demonstrations is that it allows agents to learn from expert knowledge, rather than having to discover the optimal actions through exploration. This can be particularly useful in complex or dynamic environments, where trial-and-error learning may be impractical or infeasible. For example, learning from demonstrations has been used to train agents to perform tasks such as robot manipulation \cite{saycan, rt1}, where the expert knowledge of a human operator can be used to guide the learning process.
Another advantage of leveraging demonstrations is that it can reduce the amount of data and computational resources required for training. Because the agent can learn from expert demonstrations, rather than having to discover the optimal actions through exploration, the amount of data needed to train the agent can be significantly smaller. This can make learning from demonstrations particularly attractive in situations where data collection is expensive or time-consuming.

The demonstration data can be used in numerous ways. When used offline, one can learn the policies, skills, world models, rewards, representations, or even the learning algorithm itself from the demonstrations. When used online, the demonstrations can serve as experience, regularization, reference, or curriculum. 
In Sec \ref{sec:use_demo_offline} and \ref{sec:use_demo_online}, we will go into greater detail about these various ways to use demonstrations.

It is crucial to research effective methods for collecting demonstrations of high quality and quantity to support those approaches that rely on them. Note that the demonstrations could be collected either from human experts or artificial agents (like learned policy or planners). Therefore, in Sec \ref{sec:obtain_demo} we discuss the systems that collect demonstrations with and without humans, in simulated environments and in the real world.

To summarize, we conduct a thorough analysis of numerous aspects of boosting reinforcement learning and planning through demonstrations. 
We begin by summarizing the background and preliminary knowledge about RL and planning in Sec \ref{sec:background}.
The use of demonstrations in planning and reinforcement learning is then covered in Sec \ref{sec:use_demo_offline} and \ref{sec:use_demo_online}.
Sec \ref{sec:obtain_demo} describes several existing methods of collecting demonstrations. 
In Sec \ref{sec:case_study}, a complete pipeline of collecting and utilizing demonstrations is demonstrated using ManiSkill \cite{mu2021maniskill} as an example.
Sec \ref{sec:case_study} takes ManiSkill \cite{mu2021maniskill} as an example to show a full pipeline of collecting and utilizing demonstrations. 
In Sec \ref{sec:discussion}, we conclude the entire survey and discuss the future directions in this area.

\section{Background \& Prelimiary Knowledge}

\label{sec:background}

In this section, we first define the Markov Decision Process (Sec \ref{sec:mdp}), which is the environment or task to be solved in the sequential decision making problem. And Sec \ref{sec:rl}, \ref{sec:il}, and \ref{sec:planning} briefly introduce three types of methods used to solve MDP.

\subsection{Markov Decision Process}
\label{sec:mdp}

\begin{figure}[t]
    \centering
    \includegraphics[width=0.9\linewidth]{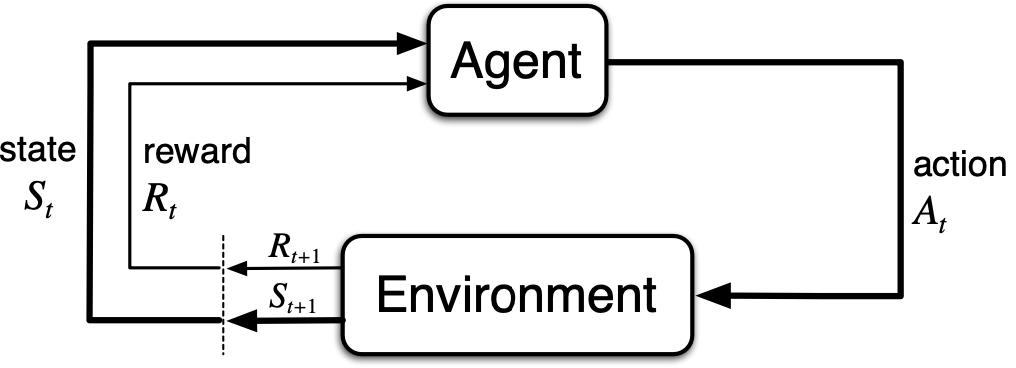}
    \caption{
        The agent–environment interaction in a Markov decision process from \cite{rl_book}
    }
    \label{fig:mdp}
\end{figure}

An MDP, denoted as $\mathcal{M}:=\langle S,A,T,R,\gamma \rangle$, serves as a model for an agent's sequential decision-making process. Here, $S$ represents a finite set of states, and $A$ is a set of actions. The mapping $T:S\times A \to \textsf{Prob}(S)$ defines a probability distribution over the set of next states, given that the agent takes action $a$ in state $s$. Here, $\textsf{Prob}(S)$ refers to the set of all probability distributions over $S$. The transition probability $T(s'|s,a)\in[0,1]$ denotes the probability that the system transitions to state $s'$.

The reward function $R$ can be specified in various ways. When $R:S \to \mathbb{R}$, it gives the scalar reinforcement at state $s$. Alternatively, $R:S \times A \to \mathbb{R}$ maps a tuple (state $s$, action $a$ taken in state $s$) to the reward received upon performing the action. Finally, $R:S \times A \times S \to \mathbb{R}$ maps a triplet (state $s$, action $a$, resultant state $s'$) to the reward obtained upon performing the transition.

The discount factor $\gamma \in [0,1]$ serves as the weight for past rewards accumulated in a trajectory $\langle (s_{0},a_{0}),(s_{1},a_{1}),\ldots,(s_{j},a_{j}) \rangle$, where $s_{j}\in S$, $a_{j}\in A$, and $j \in\mathbb{N}$. The trajectory represents a sequence of state-action pairs that the agent takes during its decision-making process.

Fig \ref{fig:mdp} shows the agent–environment interaction interface in a Markov decision process.

\subsection{Reinforcement Learning}
\label{sec:rl}

Reinforcement Learning (RL) aims to determine an optimal policy in an MDP $\mathcal{M}$ by interacting with it.

A \emph{policy} refers to a function that maps the current state to the next action choice(s). It can be deterministic, $\pi:S\to A$, or stochastic, $\pi:S\to \textsf{Prob}(A)$. For a given policy $\pi$, the value function $V^{\pi}:S \to \mathbb{R}$ provides the value of a state $s$, which is the long-term expected cumulative reward incurred by following $\pi$ from that state. Specifically, the value of a policy $\pi$ for a given start state $s_0$ is defined as follows:

\begin{equation}
V^{\pi}(s_0) = E_{s,\pi(s)}\left[ \sum_{t=0}^\infty \gamma^t
R(s_t,\pi(s_t))|s_0 \right]
\label{eqn:state_value}
\end{equation}

In RL, the goal is to find an optimal policy $\pi^*$ that satisfies $V^{\pi^*}(s)=V^*(s)=\sup_\pi~ V^{\pi}(s)$ for all $s\in S$. The action-value function for $\pi$, $Q^{\pi}: S\times A  \to \mathbb{R}$, maps a state-action pair to the long-term expected cumulative reward incurred after taking action $a$ from $s$ and following policy $\pi$ thereafter. Additionally, we define the optimal action-value function as $Q^*(s,a)=\sup_\pi  Q^{\pi}(s,a)$. Then, $V^*(s) = \sup_{a \in A} Q^*(s,a)$.

RL offers an online approach for solving an MDP. The input for RL consists of sampled experiences in the form of $(s, a, r)$ or $(s, a, r, s')$, which includes the reward or reinforcement due to the agent performing action $a$ in state $s$. In the model-free setting of RL, the transition function $T$ is unknown. Both the transition function and policy are estimated from the samples, and the goal of RL is to learn an optimal policy.

\subsection{Imitation Learning}
\label{sec:il}

Imitation learning (IL) is an alternative approach to finding good policies for an MDP $\mathcal{M}$. IL seeks to extract knowledge from demonstrations provided by experts in order to replicate their behaviors. It is often used in complex sequential decision-making problems where pure reinforcement learning would require a large number of samples to solve.

There are generally two types of imitation learning methods. Behavior cloning (BC) \cite{bc} is a simple yet effective method that learns a policy in a supervised manner by cloning the expert's actions at each state in the demonstration dataset. Another approach to imitating expert behaviors is through inverse reinforcement learning (IRL) \cite{irl, gail}, which finds a reward function that allows for the solution of the expert demonstrations.

\subsection{Planning with Models}
\label{sec:planning}
When the environment model, including the transition function $T$ and reward function $R$, is available to the agent, planning can be used to solve the MDP $\mathcal{M}$. \emph{Planning} refers to any computational process that takes a model as input and produces or improves a policy for interacting with the modeled environment. This approach is sometimes considered a type of model-based RL.

\begin{figure}[t]
    \centering
    \includegraphics[width=0.9\linewidth]{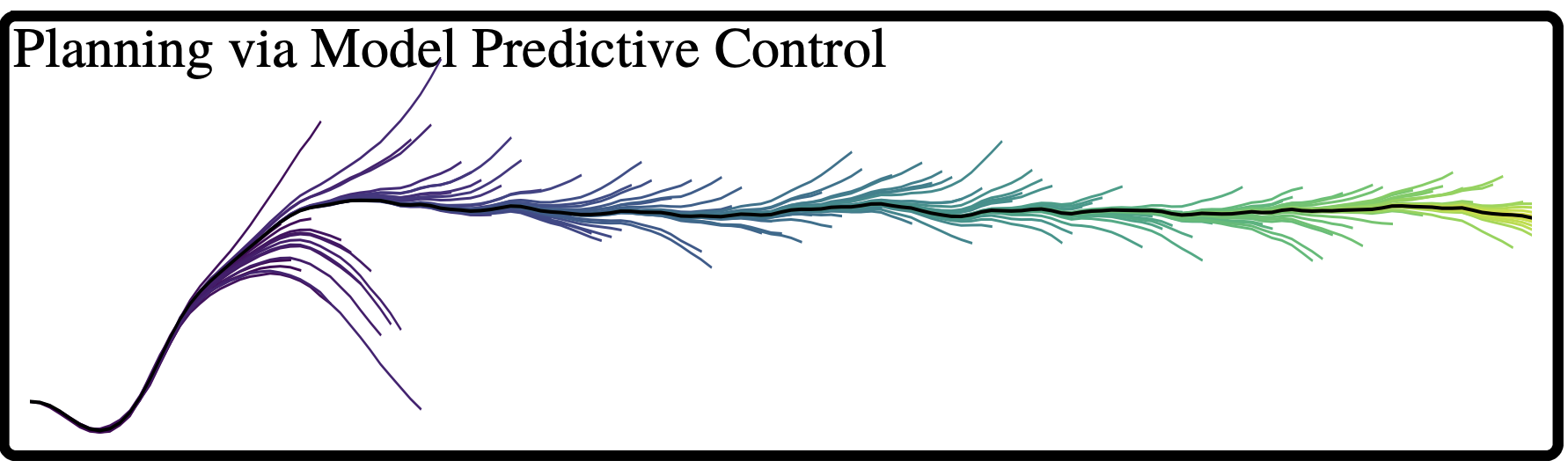}
    \caption{
        An illustration of Model Predictive Control from \cite{pets}
    }
    \label{fig:mpc}
\end{figure}

Model predictive control (MPC) \cite{mpc_book} is a model-based control method that plans an optimized sequence of actions in the model. Typically, at each time step, MPC obtains an optimal action sequence by sampling multiple sequences and applying the first action of the sequence to the environment. Formally, at time step $t$, an MPC agent seeks an action sequence $a_{t:t+\tau}$ by optimizing:

\begin{equation}
\label{eq_mpc_basic}
    \max_{a_{t:t+\tau}}~\mathbb{E}_{s_{t'+1}\sim p(s_{t'+1}|s_{t'},a_{t'})}\left[ \sum_{t'=t}^{t+\tau}r(s_{t'}, a_{t'}) \right]
\end{equation}
where $\tau$ denotes the planning horizon. Then the agent will choose the first action $a_t$ from the action sequence and apply it to the environment. An illustration of MPC is shown in Fig \ref{fig:mpc}.

Monte Carlo tree search (MCTS) \cite{mcts} is an extension of Monte Carlo sampling methods that aim to solve Eq.(\ref{eq_mpc_basic}). Unlike the Monte Carlo methods used in the MPC approach, MCTS uses a tree-search method. At each time step, MCTS incrementally extends a search tree from the current environment state. Each node in the tree corresponds to a state, which is evaluated using some approximated value functions or the return obtained after rollouts in the model with a random policy~\cite{mcts} or a neural network policy~\cite{alphago, alphago_zero, alpha_zero}. Finally, an action is chosen that is more likely to transition the agent to a state with a higher evaluated value. In MCTS, models are typically used to generate the search tree and evaluate the state.

\section{Using Demonstrations Offline}
\label{sec:use_demo_offline}

Most methods that utilize demonstrations can be divided into two stages: an offline stage and an online stage, both of which are optional. During the offline stage, the agent cannot access the environment and can only learn from the demonstrations. In the online stage, the agent is allowed to interact with the environment and can learn from both the environment and the demonstrations to boost the learning process.

In this section, we will mainly discuss how to learn from demonstrations in the offline stage. There can be a subsequent online stage to further improve the agent's performance, which we will discuss in Sec \ref{sec:use_demo_online}. Note that demonstrations can be used in both stages, but we separate them to better categorize different strategies for utilizing demonstrations.

When learning from demonstrations offline, there are various things that can be learned, such as policies, skills, world models, rewards, representations, and more. We will discuss these topics one by one.

\subsection{Learning Policy}

Learning policies directly from demonstrations is perhaps the most straightforward way to leverage the knowledge contained in the demonstrations. The basic idea behind \textit{imitation learning (IL)}, as we introduced in Sec \ref{sec:il}, is to imitate the actions in demonstrations based on observations. IL methods can be roughly divided into two types: behavior cloning (BC) and inverse reinforcement learning (IRL). Although behavior cloning may seem simple, it is widely used in practice, even for challenging tasks like real-world robot manipulation \cite{saycan, rt1}. IRL is a bit more complex because it first estimates a reward function from the demonstration data and then solves a forward RL problem using this reward function. We will discuss IRL in more detail in Sec \ref{sec:learn_reward_offline} and \ref{sec:demo_as_ref}.

One limitation of IL is that it requires expert demonstrations, meaning that the policy used to collect the demonstrations should be as optimal as possible. Since the agent only imitates the behavior in the demonstration, it cannot outperform the demonstrator. However, in many scenarios, we may only have access to sub-optimal demonstrations, such as random-play demonstrations. In this case, we need a method that can potentially learn a better policy solely from the demonstrations, and this is the goal of offline reinforcement learning.

Recently, there have been many exciting works in the area of offline RL \cite{bcq, bear, brac, rem, cql}, which focus on learning a policy using only the offline data without any online learning or online fine-tuning.

In the methods we discussed above, the policy is learned solely from the demonstrations, and therefore, it is limited by the performance of the policy that generated the demonstration data. Even with offline RL methods, the learned policy may not be very strong due to the distributional shift between the demonstrated states and the policy's own states \cite{distribution_shift}. Therefore, a common practice is to fine-tune the policies learned offline through online learning. For example, in AlphaGo \cite{alphago}, the policy network is first pre-trained by cloning human demonstrations and then fine-tuned through reinforcement learning. Similarly, \cite{awac, aw_opt} fine-tunes the policy pre-trained by offline RL.
Another approach is to leverage a pre-trained policy to generate potential action candidates during online planning, similar to the approach used by \cite{yinplanning}. 


\subsection{Skill Discovery}


In some cases, the demonstrations may not be for the exact same task that we want to solve. However, those tasks may share the same skill set. For instance, making coffee and washing a cup both require the skill "pick and place a cup." Here, we use the term "skill" to refer to some short-horizon policies shared among many tasks. It can be challenging to learn a monolithic RL policy for complex, long-horizon tasks due to issues such as high sample complexity, complicated reward design, and inefficient exploration. Therefore, a natural approach is to learn a set of skills from demonstrations and then use the learned skills in downstream tasks.


There is a line of research \cite{niekum2013incremental, fox2017multi, krishnan2017ddco, sharma2018directed, kipf2019compile, bera2019podnet, pertsch2020accelerating, zhou2020plas, zhao2020augmenting, lee2020learning, ajay2020opal, Shankar2020Discovering, shankar2020learning, lu2021learning, zhu2021bottom, rao2021learning, tanneberg2021skid, yang2021trail} that predominantly learns skills using a latent variable model, where the latent variables partition the experience into skills, and the overall model is learned by maximizing (a lower bound on) the likelihood of the experiences. This approach is structurally similar to a long line of work in hierarchical Bayesian modeling \cite{ghahramani2000variational, blei2001topic, fox2011sticky, linderman2017bayesian}.

Another approach is to use a combination of the maximum likelihood objective and a penalty on the description length of the skills, which is proposed by LOVE \cite{love}. This penalty incentivizes the skills to maximally identify and extract common structures from the experiences.

Goal-conditioned policies can also be viewed as a type of skill. For instance, \cite{franka_kitchen} learns a goal-conditioned policy from unstructured demonstrations by goal relabeling \cite{andrychowicz2017hindsight}. This goal-conditioned policy provides a good policy initialization for subsequent hierarchical reinforcement fine-tuning.

To use the learned skills to boost online reinforcement learning, Parrot \cite{singh2020parrot} provides a concrete example. Parrot first learns a behavioral prior from a multi-task demonstration dataset, which is essentially an invertible mapping that maps noise to useful action. When training to solve a new task, instead of directly executing its actions in the original MDP, the agent learns a policy that outputs $z$, which is taken by the behavioral prior as input. Then, the output from the behavioral prior is executed in the environment.


\subsection{Learning World Model}
\label{sec:learn_model}

In the methods we discussed earlier, the components learned from demonstrations, such as policies and skills, are usually fine-tuned online using model-free reinforcement learning algorithms. However, in model-based RL, which is a crucial category of RL algorithms, the world model is an essential component. As noted in the previous section, the world model can be represented as the MDP $\langle S, A, T, R, \gamma \rangle$, where $S$, $A$, and $\gamma$ are typically predefined, and the state transition dynamics $T$ and reward function $R$ must be learned. Since the reward function can be learned in a similar manner to the transition function, we will focus on the learning of the transition function for simplicity.

When given a demonstration dataset, one straightforward approach to learning the world model is to fit the one-step transitions. If $T_\theta$ is deterministic, the model learning objective can be the mean squared prediction error of $T_\theta$ on the next state.

\begin{equation}
    \min_{\theta} \mathbb{E}_{(s, a) \sim D, s^\prime \sim T^* (\cdot|s, a)}  \left[ \left\Vert s^\prime - T_\theta (s, a) \right\Vert_{2}^2 \right].
    \label{eq:model_objective_mse}
\end{equation}

Here $T^*$ is the real transition dynamics, and $D$ is the demonstration dataset. 

However, deterministic transition models fail to capture the aleatoric uncertainty \cite{pets} that arises from the inherent stochasticities of the environment. To model this uncertainty, a natural approach is to use a probabilistic transition model $T_\theta (\cdot|s, a)$ \cite{pets}. In this case, the objective can be to minimize the KL divergence between the true transition distribution $T^* (\cdot|s, a)$ and the learned distribution $T_\theta (\cdot|s, a)$, as shown in Eq. \eqref{eq:model_objective_kl}.
\begin{equation}
\begin{split}
\label{eq:model_objective_kl}
    \min_{\theta} \mathbb{E}_{(s, a) \sim D} \left[D_{\mathrm{KL}}\left(T^*(\cdot | s, a), T_{\theta}(\cdot | s, a)\right)\right] := \\
    \mathbb{E}_{(s, a) \sim D, s^\prime \sim T^* (\cdot|s, a)} \ls \log \lp \frac{T^*\left(s^{\prime} \mid s, a\right)}{T_{\theta}\left(s^{\prime} \mid s, a\right)} \rp \rs .
\end{split}
\end{equation}

When learning the world model, the requirements of the demonstration dataset are different from those used for learning policy. A good world model should cover the most possible states, so the trajectories in the demonstration dataset should be as diverse as possible. Furthermore, the actions in the demonstrations do not necessarily have to be optimal. In other words, the demonstration dataset should provide a diverse set of experiences.

After the world model is learned offline, it can be used to solve the MDP in two different ways.

The first way is to use it as a proxy of the true MDP and run any RL algorithm within this learned model. If the agent is allowed to interact with the true MDP, the collected data can be used to further improve the learned world model. This strategy is adopted by several works such as \cite{me_trpo, dreamer, mbrl_atari}.

The second way is to plan in it. For example, \cite{planet} runs MPC in a learned world model to perform vision-based continuous control, and \cite{muzero} uses MCTS with the learned model to play various video games.

However, directly planning in the world model learned from demonstrations can have arbitrarily large sub-optimality, as demonstrated by \cite{ross2012agnostic}, because the demonstration dataset may not span the entire state space, and the learned world model may not be globally accurate.

One simple solution is to fine-tune the world model by interacting with the real MDP, which is also a common practice in model-based RL. Recently, more sophisticated methods have been proposed to address this issue. For instance, MOReL \cite{kidambi2020morel} constructs a pessimistic MDP (P-MDP) using the learned model and an unknown state-action pair detector, which penalizes policies that venture into unknown parts of state-action space. Another approach is taken by \cite{yu2020mopo}, which learns a world model from the demonstration dataset along with an uncertainty estimator $u(s,a)$ of the world model. This uncertainty estimator is then used to penalize the reward function.


\subsection{Inferring Reward}
\label{sec:learn_reward_offline}

In some scenarios, the reward function is not present in the demonstrations, making it challenging to learn the reward as part of the world model. However, it is still possible to infer the reward function if the trajectories have sparse labels, such as success and failure. Note that the methods described in this section can be regarded as inverse RL methods, but they differ from the most common inverse RL methods, which typically involve interaction with the environment. Online IRL methods will be discussed in Sec \ref{sec:demo_as_ref}.

One approach to inferring the reward function from demonstrations is to learn a goal classifier from the demonstrations and use it as the reward function \cite{xie2018few, vecerik2019practical, singh2019end}. Specifically, the goal observations (usually in the form of images) can be extracted from successful trajectories, and the negative samples can be extracted from failed trajectories. The reward function (goal classifier) can then be trained with a binary cross-entropy loss and used in online reinforcement learning problems where the reward function is not presented.

Another approach that leverages almost all data samples is proposed by \cite{zolna2020offline}. This method first learns a reward function by contrasting observations from the demonstrator and unlabeled trajectories, annotates all data with the learned reward, and finally trains an agent via offline reinforcement learning.


\subsection{Learning Representations}

Representation learning via supervised / self-supervised / unsupervised pre-training on large-scale datasets has emerged as a powerful paradigm in computer vision \cite{simclr, moco, doersch2015unsupervised, oord2018representation} and natural language processing \cite{devlin2018bert, brown2020language, chowdhery2022palm}, where large datasets are available. When a large demonstration dataset is available, it is natural to apply representation learning to improve policy learning. In this case, the demonstration dataset does not need to be in the form of trajectories since the observation alone can be sufficient for representation learning.

The intuition behind using representation learning to improve policy learning is straightforward. Standard visual policy learning frameworks attempt to simultaneously learn a succinct but effective representation from diverse visual data while also learning to associate actions with such representations. Such joint learning often requires large amounts of demonstrations to be effective. Decoupling the two problems by first learning a visual representation encoder from offline data using standard supervised and self-supervised learning methods and then training the policy based on the learned, frozen representations can be an effective approach.

Empirically, VINN \cite{pari2021surprising} demonstrates that this simple decoupling improves the performance of visual imitation models on both offline demonstration datasets and real-robot door-opening tasks. Similarly, \cite{yang2021representation} shows that pre-training with unsupervised learning objectives can significantly enhance the performance of policy learning algorithms that otherwise yield mediocre performance.

Recently, there has been rapid progress in the development of algorithms for representation learning, but the importance of data in representation learning for sequential decision making cannot be overstated. To this end, \cite{zhan2022learning} has proposed collecting an in-domain dataset and pre-training the visual encoder on it. An in-domain dataset is one that is obtained from the same domain as the tasks to be solved. Conversely, several works have explored learning policies with visual representations pre-trained on large external datasets \cite{shah2021rrl, pari2021surprising, nair2022r3m, wang2022vrl3, mvp2}. However, it remains unclear whether one approach is inherently superior to the other.


\subsection{Learning Trajectory Imitator}

\begin{figure}[t]
    \centering
    \includegraphics[width=0.9\linewidth]{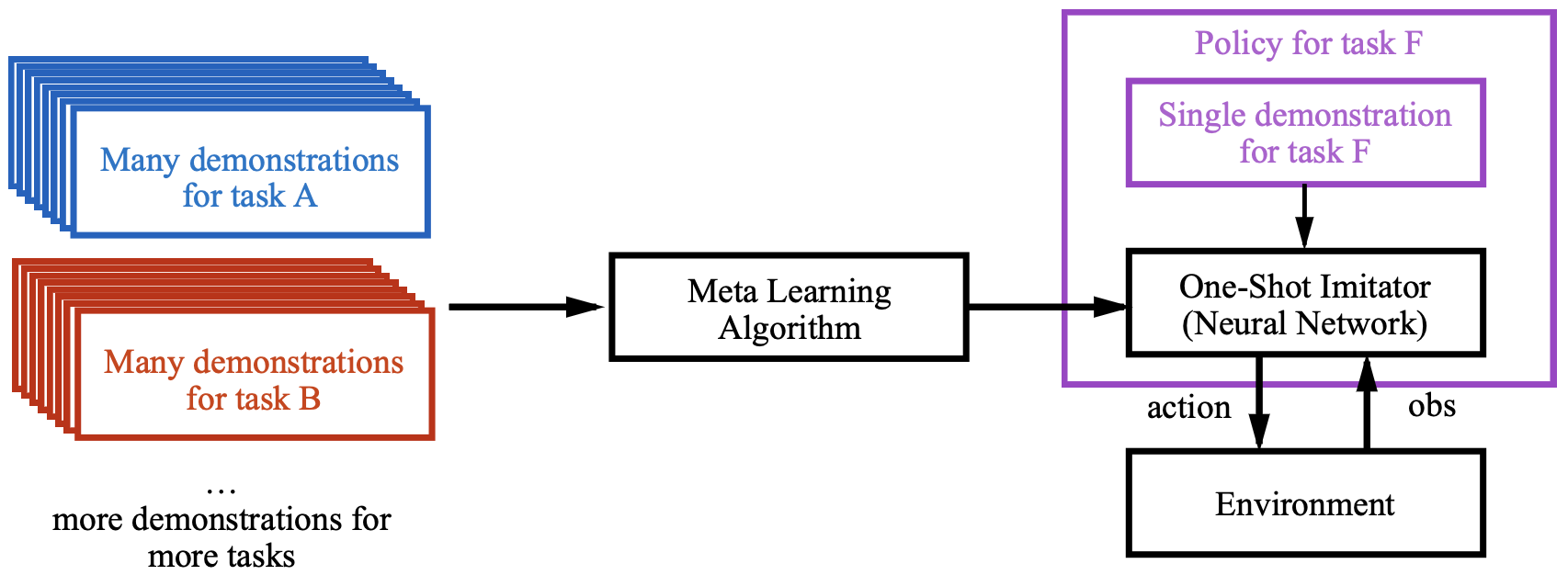}
    \caption{
        An illustration of One-Shot Imitation Learning from \cite{one_shot_il}
    }
    \label{fig:one_shot_il}
\end{figure}

Imitation learning is a commonly used technique for solving different tasks, but each new task requires separate training of the agent. Ideally, agents should be able to perform a novel task by imitating a demonstration trajectory given at test time. This is the \textit{one-shot imitation learning} problem proposed by \cite{one_shot_il}. We refer to the agent with such capability as a trajectory imitator, which can also be learned from demonstrations offline.

Given multiple demonstrations from \textit{different tasks}, the one-shot imitation learning problem can be formulated as a supervised learning problem. To train the trajectory imitator, in each iteration a demonstration is sampled from one of the training tasks and fed to the network. Then, a state-action pair is sampled from a second demonstration of the same task. The network is trained to output the corresponding action when conditioned on both the first demonstration and this state. An illustration of the problem setting is shown in Fig \ref{fig:one_shot_il}. This framework has been extended to visual imitation \cite{finn2017one, pathak2018zero}, cross-domain imitation \cite{yu2018one}, language-conditioned imitation \cite{lynch2020language}, and more.

The process of providing a demonstration trajectory at test time is similar to the prompting process \cite{wei2022chain} used in large language models. Recently, Gato \cite{gato} combined the idea of one-shot imitation learning with the large language model to build a single generalist agent beyond the realm of text outputs. The agent works as a multi-modal, multi-task, multi-embodiment generalist policy. It can play Atari, caption images, chat, stack blocks with a real robot arm, and much more, deciding based on its context whether to output text, joint torques, button presses, or other tokens.


\subsection{Learning Algorithm}

\begin{figure}[t!]
    \centering
    \includegraphics[width=1.0\linewidth]{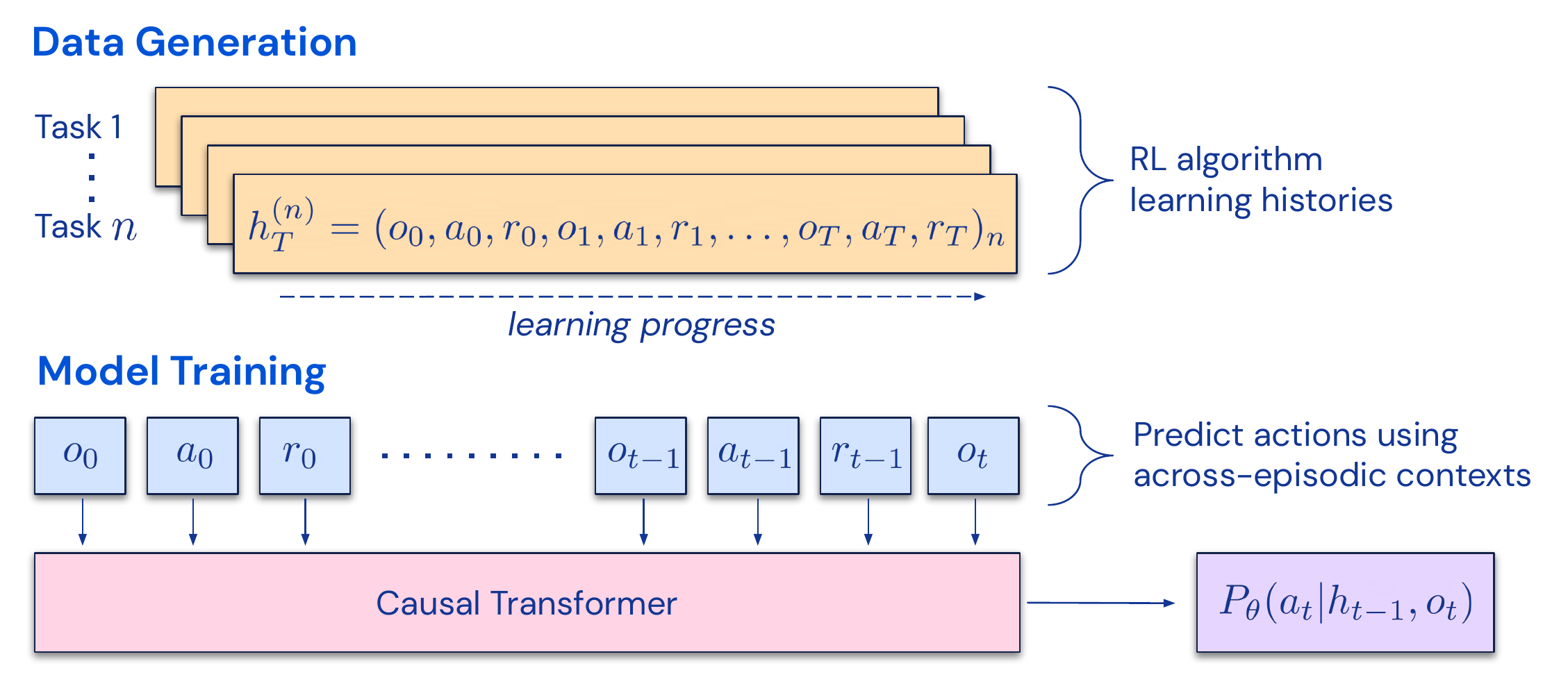}
    \vspace{-0.7cm}
    \caption{
        An illustration of Algorithm Distillation from \cite{laskin2022context}
    }
    \label{fig:algo_distillation}
\end{figure}

Assuming the demonstration dataset does not only contain the expert demonstration trajectories, but also the entire \textit{learning history} of an agent, e.g., the whole replay buffer of an RL agent, is it possible to learn the "learning progress" from the dataset?

Algorithm Distillation (AD) \cite{laskin2022context} is a recent approach that allows us to learn reinforcement learning algorithms directly from offline datasets, including the complete learning histories of agents. AD involves two steps. First, we train multiple instances of an RL algorithm to solve a variety of tasks and save their learning histories. Then, using this dataset of learning histories, we train a transformer model to predict actions given the preceding learning history. Since the policy improves over time, accurately predicting actions requires the transformer to model policy improvement. The resulting transformer can explore, exploit, and maximize return when its weights are frozen, allowing it to learn the reinforcement learning algorithm itself. Figure \ref{fig:algo_distillation} provides a visual illustration of the entire algorithm distillation pipeline.
\section{Using Demonstrations Online}
\label{sec:use_demo_online}

In addition to offline learning from demonstrations, it is also possible to utilize demonstration data directly during online learning, without the need for a preceding offline learning stage. In this section, we will discuss the different methods of online reinforcement learning or planning that can be enhanced through the use of demonstrations.

\subsection{Demo as Off-Policy Experience}

One approach to incorporating demonstration data directly into online learning is to integrate it with modern off-policy reinforcement learning methods \cite{dqn, rainbow, ddpg, td3, sac}, which maintain a replay buffer \cite{dqn} and optimize the policy by training on the experience sampled from the replay buffer. In this approach, demonstration trajectories are simply added to the replay buffer, and off-policy reinforcement learning is run as usual. The demonstration data is never overwritten in the replay buffer, ensuring that it remains available throughout the training process. This strategy, which is simple and easy to implement, has been widely adopted by methods such as \cite{dqnfd,r2d2,r2d3,ddpgfd,nair2018overcoming}. By including the demonstration data in the replay buffer, the agent is exposed to a more diverse set of states before exploring them, addressing the exploration problem, which is a central challenge in reinforcement learning.

Despite its potential advantages, the aforementioned strategy has certain limitations that prevent it from effectively exploiting the demonstration data. Firstly, the approach relies solely on expert trajectories as learning references and may fail to fully utilize their effect when the number of demonstrations is limited, as confirmed by \cite{pofd}. Secondly, the demonstrated trajectories must be associated with rewards for each state transition to be compatible with collected data during training. However, the rewards used in the demonstrations may differ from those used to learn the policy in the current environment \cite{max_entropy_irl}, or they may be unavailable.

In situations where rewards are absent from the demonstrations, the SQIL algorithm \cite{sqil} can still be applied. This approach adds the demonstrations to the replay buffer with a constant reward of 1, while all other experiences collected online are added to the buffer with a constant reward of 0. This can be viewed as a regularized variant of behavior cloning that employs a sparsity prior to encourage long-horizon imitation.

\subsection{Demo as On-Policy Regularization}
\label{sec:demo_as_reg}

Off-policy methods have the potential to be more sample efficient, but are often plagued with instability issues \cite{duan2016benchmarking, drl_matters}. In contrast, on-policy methods tend to be more stable and perform well in high-dimensional spaces \cite{duan2016benchmarking}. Consequently, several studies have explored ways to incorporate demonstrations into on-policy reinforcement learning algorithms.

\begin{figure}[t]
    \centering
    \includegraphics[width=0.9\linewidth]{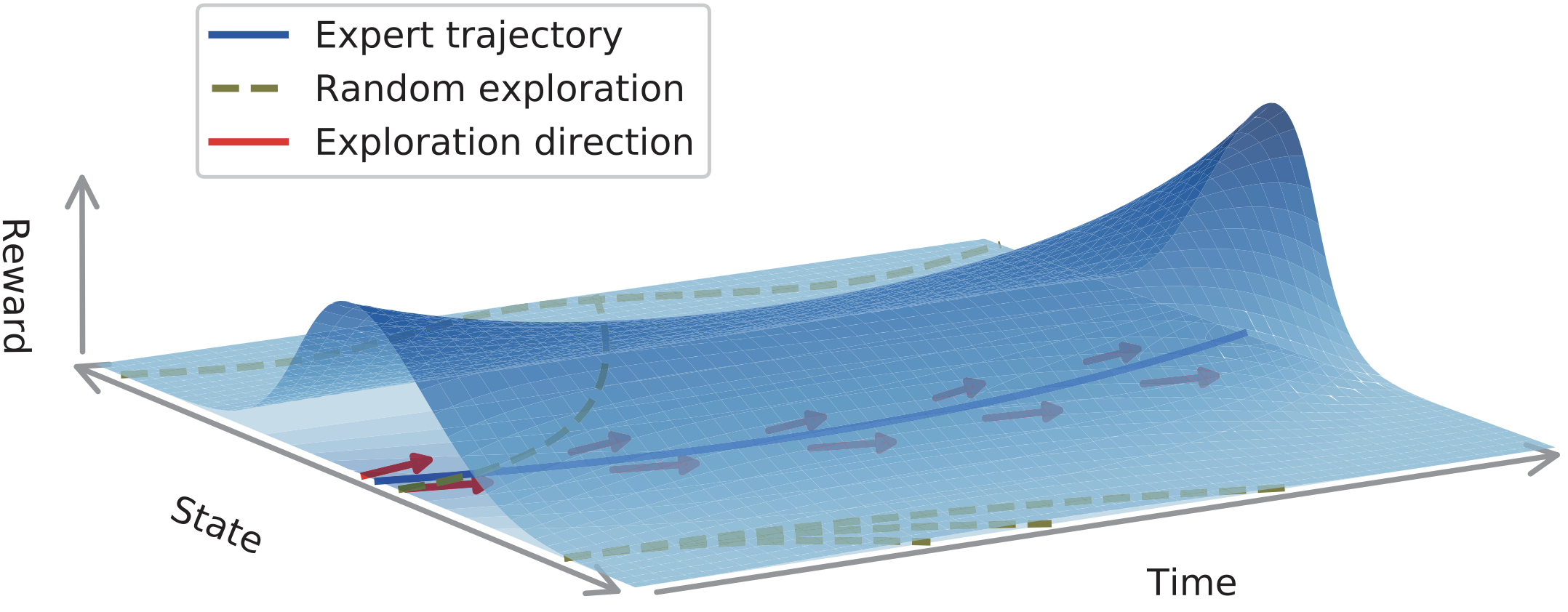}
    \caption{
        POfD \cite{pofd} explores in the high-reward regions (red arrows), with the aid of demonstrations (the blue curve). It thus performs better than random explorations (olive green dashed curves) in sparse-reward environments.
    }
    \label{fig:pofd}
\end{figure}

In POfD \cite{pofd}, the authors propose using expert demonstrations to guide exploration in reinforcement learning by enforcing occupancy measure matching between the learned policy and the demonstrations. This is achieved by introducing a demonstration-guided exploration term to the vanilla RL objective $\eta(\pi_\theta)$:
\begin{equation}
\mathcal{L}\left(\pi_\theta\right)=-\eta\left(\pi_\theta\right)+\lambda_1 D_{J S}\left(\rho_\theta, \rho_E\right)
\end{equation}
Here, $D_{JS}$ denotes the Jensen-Shannon divergence, while $\rho_\theta$ and $\rho_E$ represent the occupancy measures of the agent policy and the expert policy, respectively. By adding this regularization term, the agent is encouraged to explore areas demonstrated by the expert policy, as depicted in Figure \ref{fig:pofd}. This approach is compatible with a variety of policy gradient methods, such as TRPO \cite{trpo} and PPO \cite{ppo}.

DAPG \cite{dapg} first pre-trains the policy using behavior cloning to provide an informed initialization that can efficiently guide exploration. During the online RL fine-tuning phase, DAPG adds an additional term to the policy gradient to make full use of the information in the demonstration data:
\begin{equation}
\begin{split}
    g_{aug}=\sum_{(s,a)\in\rho_\pi} \nabla_\theta\ln\pi_\theta(a|s)A^\pi(s,a) +  \\
    \sum_{(s,a)\in\rho_D} \nabla_\theta\ln\pi_\theta(a|s)w(s,a)
\end{split}
\end{equation}

Here, $\rho_\pi$ represents the dataset obtained by executing policy $\pi$ on the MDP, and $w(s,a)$ is a weighting function. The first term is the gradient from the RL algorithm, and the second term is the gradient from the behavior cloning loss. In practice, $w(s,a)$ is implemented using a heuristic weighting function:
\begin{equation}
    w(s, a)=\lambda_0 \lambda_1^k \max _{\left(s^{\prime}, a^{\prime}\right) \in \rho_\pi} A^\pi\left(s^{\prime}, a^{\prime}\right) \quad \forall(s, a) \in \rho_D
\end{equation}

where $\lambda_0$ and $\lambda_1$ are hyperparameters, and $k$ is the iteration counter. The decay of the weighting term via $\lambda_1^k$ is motivated by the premise that, initially, the actions suggested by the demonstrations are at least as good as the actions produced by the policy. However, towards the end, when the policy is comparable in performance to the demonstrations, we do not want to bias the gradient. Thus, the auxiliary objective asymptotically decays.

\subsection{Demo as Reference for Reward}
\label{sec:demo_as_ref}

In cases where the reward is sparse or unavailable, demonstrations can be used to compute the reward. This can be accomplished by either directly defining a reward based on the demonstration or by matching the agent's trajectory distribution with the demonstration distribution. The latter is a common strategy employed in inverse reinforcement learning. We discuss both of these methods in the following subsections.

\subsubsection{Directly Define Reward based on Demo}

If only a single (or a few) demonstration trajectories are available, a reward function can be manually designed to incentivize the agent to follow the demonstration trajectory. This is typically accomplished by defining the reward as some kind of distance between the current state and the demonstration trajectory.

Various approaches can be used to specify the reward/distance function. For example, in \cite{aytar2018playing}, a sequence of important checkpoints is extracted from the demonstration trajectory, and the agent is rewarded if it visits the checkpoints in a soft order. A checkpoint is considered visited when the distance to it is less than a threshold, and the distance is defined as cosine similarity in a learned latent space. Similarly, AVID \cite{smith2019avid} divides a demonstration trajectory into stages and trains a classifier to predict whether a state belongs to a specific stage. These classifiers can then be combined to form a reward function.

In \cite{brys2015reinforcement}, the potential of a state is first defined as the highest similarity with the states in the demonstration, and then the $\gamma$-difference of the potential function is used to construct a potential-based reward function \cite{ng1999policy}. In \cite{wu2021learning} and \cite{graph_irl}, the reward is defined as the distance to the goal in the learned latent space, where the goal is chosen as the final state in the demonstration trajectory. Recent works on motion imitation \cite{peng2018deepmimic, peng2020learning, peng2022ase} carefully design the reward as a weighted distance to the reference state, taking joint orientations, joint velocities, end effectors, and centers of mass into account.

\subsubsection{Match the Distribution of Demo}

In addition to defining a reward based on demonstrations, another approach involves allowing the agent to interact with the environment and attempting to match the distributions of agent and demonstration trajectories. During this process, the distance or divergence between agent and demonstration trajectories can be converted to a reward and used to improve the agent policy. This approach forms the basis of most inverse reinforcement learning (IRL) methods. The process can be summarized using the algorithm template shown in Algorithm \ref{alg:IRL_template} (from \cite{irl_survey}).

\begin{algorithm}[ht!]
    \caption{Template for Typical Inverse RL \cite{irl_survey}}	
    \label{alg:IRL_template}
    
    \textbf{Input:} $\mathcal{M}\backslash_{R_E}$= $ \langle S,A,T,\gamma \rangle $, Set   of   trajectories  demonstrating   desired   behavior:
        $\mathcal{D}=\{\langle
            (s_0,a_0),(s_1,a_1),\ldots,(s_t,a_t)  \rangle, \ldots  \}$,  $s_t\in S$,
            $a_t\in A$,  
        $t \in \mathbb{N}$, 
        or expert's policy: $\pi_E$, and reward function features \\
    \textbf{Output:}   $\hat{R}_E$
    \begin{algorithmic}[1]
        \State Model the expert's observed  behavior as the solution of an MDP whose reward  function is not known\; 
        \State Initialize  the parameterized  form of  the reward  function using any  given features (linearly weighted  sum of feature values, distribution over rewards, or other)\;
        \State Solve  the MDP with  the current reward function  to generate the learned behavior or policy\; 
        \State Update  the optimization parameters  to minimize  the divergence  between the  observed behavior (or policy)  and the  learned behavior (policy)\; 
        \State Repeat the previous two steps till the divergence is reduced to the desired level. 
    \end{algorithmic}
\end{algorithm}

As shown in Algorithm \ref{alg:IRL_template}, IRL methods typically involve an iterative process that alternates between reward estimation and RL, which can result in poor sample efficiency. Earlier IRL methods \cite{algo_irl, abbeel2004apprenticeship, max_entropy_irl} typically required multiple calls to a Markov decision process solver \cite{mdp_book}.

Recently, adversarial imitation learning (AIL) approaches have been proposed \cite{gail, dac, airl, ghasemipour2020divergence} that interleave the learning of the reward function with the learning process of the agent.

AIL methods operate similarly to generative adversarial networks (GANs) \cite{gan}. In these methods, a generator (the policy) is trained to maximize the confusion of a discriminator (the reward) that is itself trained to differentiate between the agent's state-action pairs and those of the expert. Adversarial IL methods essentially boil down to a distribution matching problem, i.e., the problem of minimizing the distance between probability distributions in a metric space. According to the analysis in \cite{ghasemipour2020divergence}, these methods implicitly minimize an $f$-divergence between the state-action distribution of an expert and that of the learning agent.

In addition, the POfD algorithm \cite{pofd}, mentioned in Section \ref{sec:demo_as_reg}, can also be regarded as a variant of AIL. In practice, POfD is implemented by adding a shaped reward to the environment's original reward, and the shaped reward is also computed from a discriminator that differentiates between the agent's and expert's behaviors.

Besides adversarial imitation learning (AIL) methods, there are also imitation learning-based methods that approach the problem as a distribution matching problem but do not use adversarial approaches. PWIL \cite{pwil}, for example, matches the demonstration distribution by minimizing the primal form of the Wasserstein distance, also known as the earth mover's distance. An upper bound on the Wasserstein distance can be used to compute the reward offline, avoiding the computationally expensive min-max optimization problem encountered in AIL.

\subsection{Demo as Curriculum of Start States}

When the environment is a simulator that can be fully controlled, demonstrations can be used to construct a curriculum of start states. This approach is a highly efficient exploration strategy in RL and has been studied in several prior works.

Typically, these works assume the ability to reset the simulator to arbitrary states. Therefore, they can modify the distribution of start states based on the demonstrations to simplify exploration. Some methods, such as \cite{hosu2016playing, nair2018overcoming, peng2018deepmimic}, uniformly sample states from the demonstrations as start states, while others generate curriculums of start states \cite{salimans2018learning, zhu2018reinforcement}.

Using a curriculum of start states is helpful because it reduces the exploration difficulty and accelerates learning. For example, when attempting a backflip, starting from a standing pose requires learning the early phases before progressing towards the later phases. However, if start states are sampled from the demonstrations, the agent encounters desirable states earlier and can learn faster.

\section{Demonstration Collection}
\label{sec:obtain_demo}

In this section, we discuss how to collect demonstrations. 
Particularly, we focus on demonstration collection for embodied AI (or learning-based robotics) tasks, because it involves demonstrations from both artificial agents and humans, as well as from both simulation environments and the real world. We primarily discuss how to acquire expert (or near-expert) demonstrations, as suboptimal demonstrations are relatively easy to obtain.

In demonstrations for embodied AI tasks (e.g., object manipulation), the \textit{embodiment} of the agent to complete the task can be either robot or human, and the \textit{operator} of the agent can be either an autonomous system or human. We, therefore, categorize the demonstration collection methods by these two perspectives.

\subsection{Robot Embodiment - Human Operator}

\begin{figure}[t]
    \centering
    \includegraphics[width=0.9\linewidth]{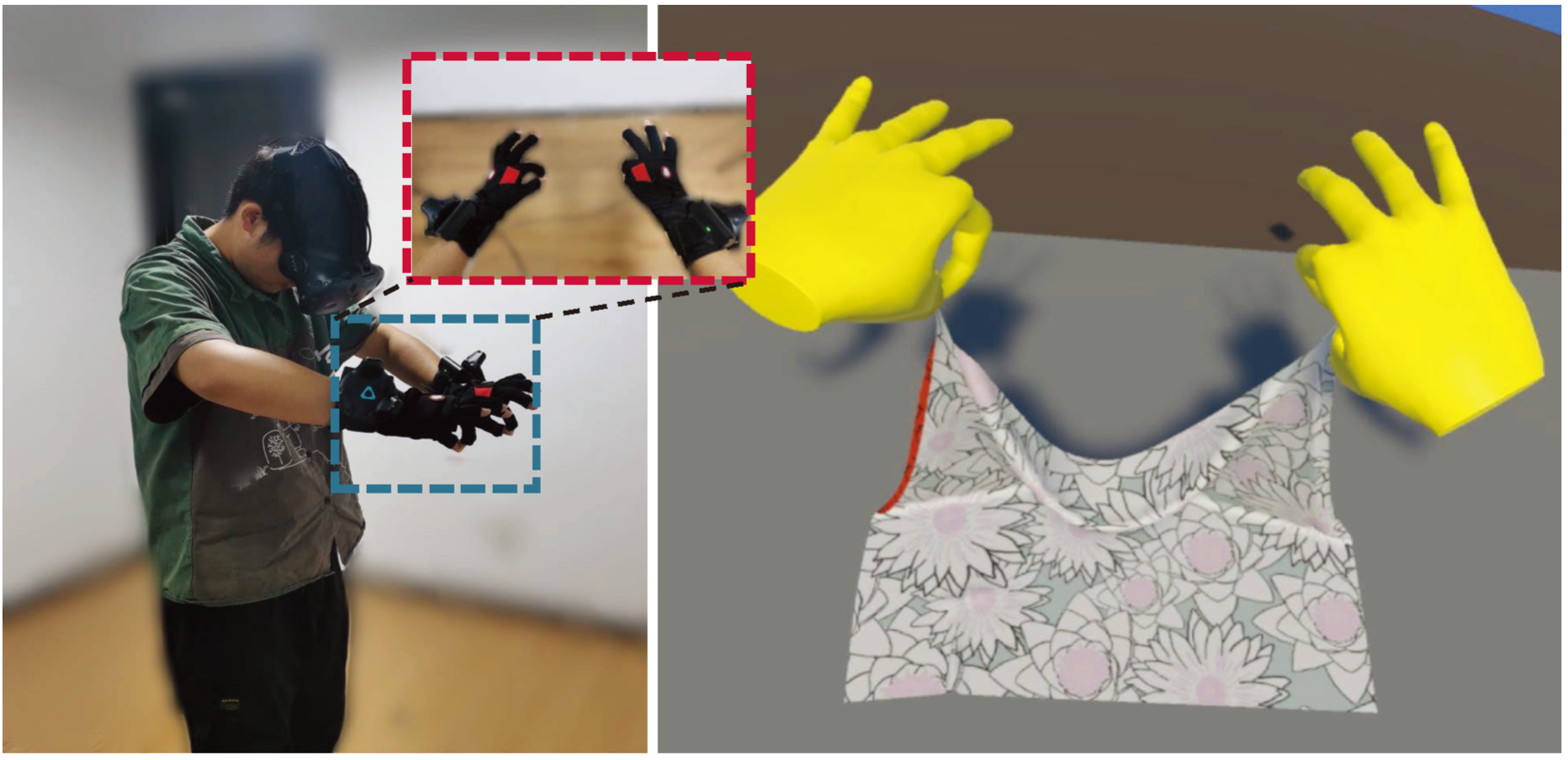}
    \caption{
       The VR interface with HTC Vive and Noitom Glove in RFUniverse \cite{rfuniverse}.
    }
    \label{fig:rfu_vr}
\end{figure}

\subsubsection{Simulation}

In the context of Embodied AI, demonstrations are often collected from human operators via teleoperation. This can be achieved using simple devices such as a mouse, keyboard, or smartphone. As these devices are easily accessible, many robot simulation environments provide interfaces that allow users to control the robot in the simulation. For example, robosuite \cite{zhu2020robosuite} provides utilities for collecting human demonstrations using a keyboard or 3D mouse devices, while iGibson \cite{igibson} is equipped with a graphical user interface that facilitates user interactions with the simulation environment. RoboTurk \cite{roboturk} offers intuitive 6 degrees-of-freedom motion control, which maps smartphone movement to robot arm movement. The simplicity of these devices makes it possible to leverage crowdsourcing to obtain data for various robotic tasks.

Recently, virtual reality (VR) interfaces have gained popularity for collecting robot demonstrations due to their flexibility. For instance, the demonstrations used in DAPG \cite{dapg} were collected in VR. Franka Kitchen environment \cite{franka_kitchen} collects demonstrations using the PUPPET MuJoCo VR system \cite{mujoco_vr}. iGibson 2.0 \cite{igibson2} includes a novel VR interface that is compatible with major commercially available VR headsets through OpenVR \cite{openvr}. RFUniverse \cite{rfuniverse} can connect to different VR devices through SteamVR, including novel VR gloves, as illustrated in Fig. \ref{fig:rfu_vr}.

\begin{figure}[t!]
    \centering
    \includegraphics[width=0.9\linewidth]{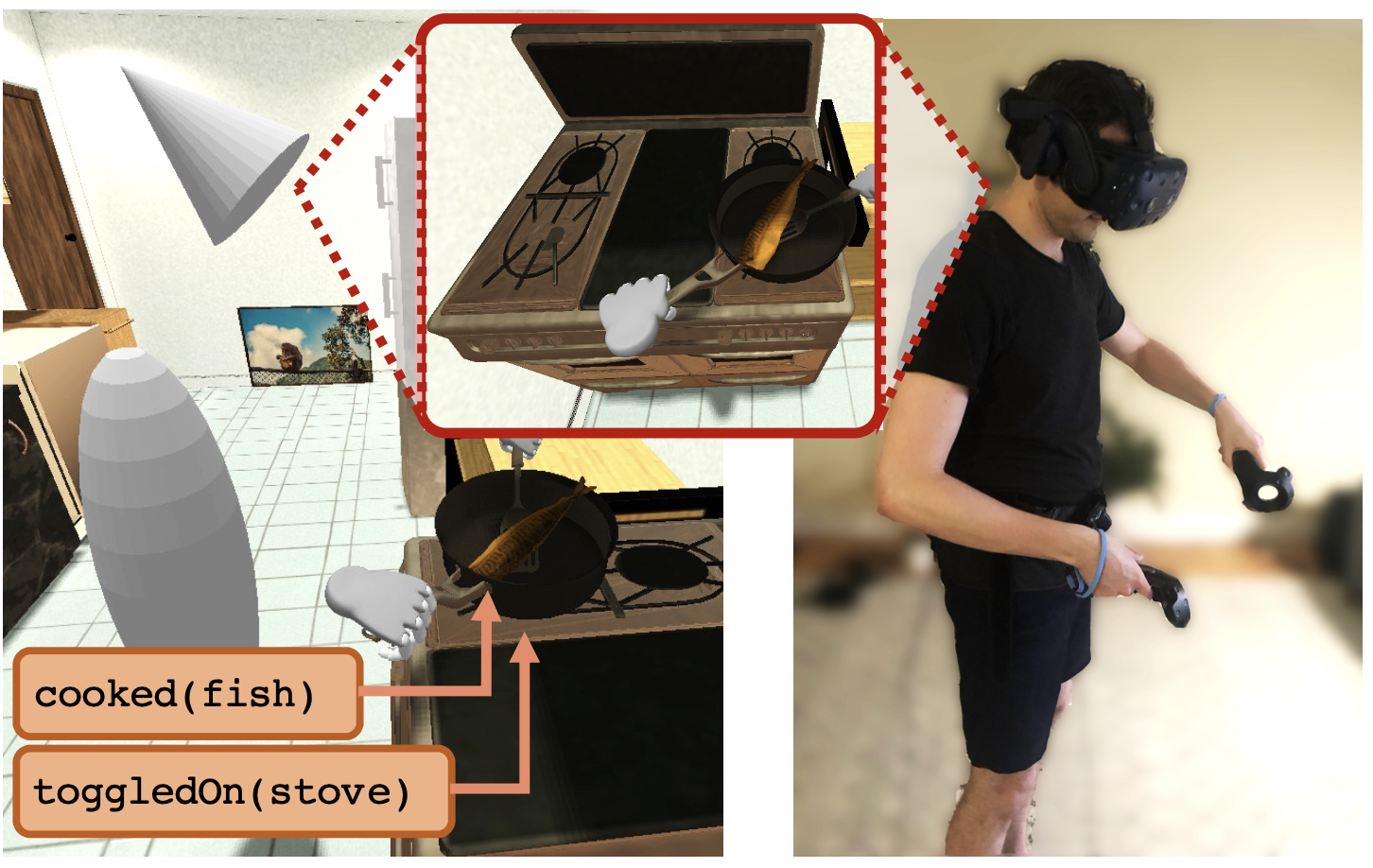}
    \caption{
       The VR interface in iGibson 2.0 \cite{igibson2}.
    }
    \label{fig:igibson2_vr}
\end{figure}

One of the major advantages of VR is the immersive experience: humans can embody an avatar in the same scene and for the same task as the AI agents. This allows for a more realistic interaction between humans and robots. For example, as shown in Fig \ref{fig:igibson2_vr}, the virtual reality avatar in iGibson 2.0 is composed of a main body, two hands, and a head. The human controls the motion of the head and the two hands via the VR headset and hand controllers, with an optional additional tracker for control of the main body. Humans receive stereo images as generated from the point of view of the head of the virtual avatar.

\begin{figure*}[t!]
    \centering
    \includegraphics[width=0.95\linewidth]{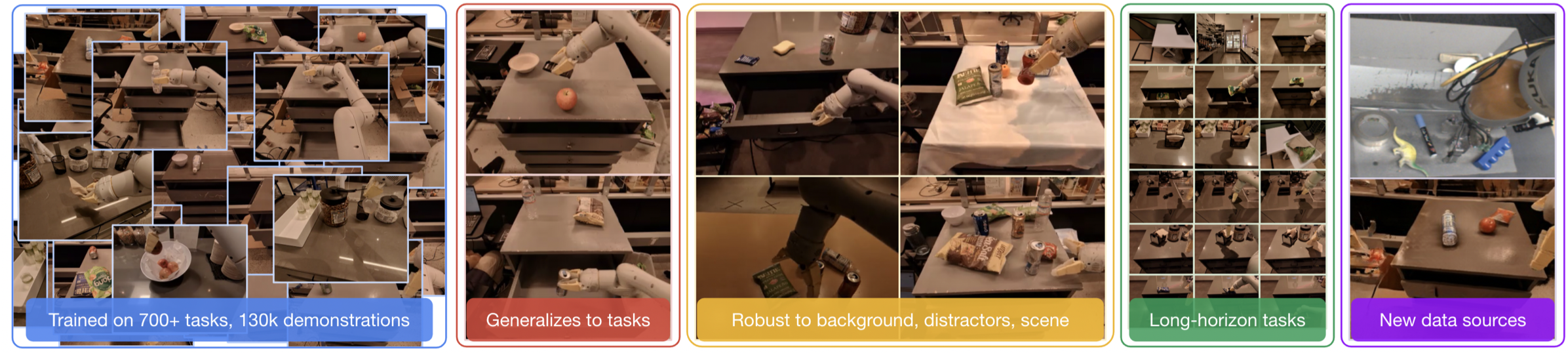}
    \caption{
       RT-1’s \cite{rt1} large-scale, real-world training (130k demonstrations) and evaluation (3000 trials) dataset.
    }
    \label{fig:rt1_data}
\end{figure*}

\subsubsection{Real World}

Furthermore, VR-based teleoperation has also been used to collect demonstrations in the real world. The SayCan system \cite{saycan} is an example of such an approach, where 68,000 teleoperated demonstrations were performed by 10 robots over 11 months. To collect the demonstrations, the operators used VR headset controllers to track the motion of their hand, which was then mapped onto the robot’s end-effector pose. In addition, a joystick was used to move the robot’s base. Human raters filtered the data to exclude unsafe, undesirable, or infeasible episodes.

With sufficient resources, VR-based teleoperation can be used to collect even larger demonstration datasets. For instance, the multi-task robot transformer RT-1 \cite{rt1} is trained on a large-scale real-world robotics dataset of 130k episodes covering over 700 tasks, which was collected using a fleet of 13 robots over 17 months. The demonstration collection process involves each of the robots autonomously approaching its station at the beginning of the episode and communicating the instruction that the operator should demonstrate to the robot. Demonstrations are collected with direct line-of-sight between the operator and robot using two virtual reality remotes, with 3D position and rotational displacements of the remote mapped to 6D displacements of the robot tool. The joystick's X and Y position is mapped to the turning angle and driving distance of the mobile base, respectively. An illustration of their dataset is shown in Fig \ref{fig:rt1_data}.

\subsection{Robot Embodiment - Autonomous Operator}

\subsubsection{Simulation}

While collecting demonstrations by humans is a straightforward approach, scalability is still a potential issue. To address this, a number of works have explored the use of autonomous agents to generate demonstrations in a more scalable manner.

One approach to generate large-scale demonstrations is by using a \textit{planner}. In some high-level planning tasks, such as Watch-And-Help \cite{puig2020watch} and ALFRED \cite{shridhar2020alfred}, symbolic planners \cite{hoffmann2001ff} can be used to generate demonstrations at a low cost. Meanwhile, in low-level control tasks, such as robot manipulation, motion planners \cite{kuffner2000rrt, prm} can be utilized. For example, RLbench \cite{rlbench} relies on an infinite supply of generated demonstrations collected via motion planners. Compared to the crowd-sourcing way to collect demonstrations, the planner-based system cannot be run in the real world, but in exchange, it receives the ability to generate a diverse range of tasks in a scalable way.

In addition to planning, learning-based methods can also be used to generate demonstrations. Various different methods are deployed in D4RL \cite{d4rl}. For Gym-MuJuCo tasks, expert demonstrations are generated by RL agents. For AntMaze tasks, they are generated by training a goal-reaching policy and using it in conjunction with a high-level waypoint generator. In Adroit, a large amount of expert data can be obtained by a learning-from-demonstration agent trained with a small number of human demonstrations.

ManiSkill \cite{mu2021maniskill} also uses RL agents to collect demonstrations, but training a single RL agent to solve a task can be challenging due to the difficulty of the tasks. Therefore, it trains RL agents in a divide-and-conquer way, which is a more scalable way to solve tasks and generate demonstrations. This will be discussed in detail in Section \ref{ms_demo_collection}.

\subsubsection{Real World}

Collecting demonstration data in the real world can be challenging due to the high cost of sample collection and the limitations of some methods, such as motion planners. To address these challenges, demonstration collection systems must be designed carefully.

\begin{figure}[t]
    \centering
    \includegraphics[width=0.9\linewidth]{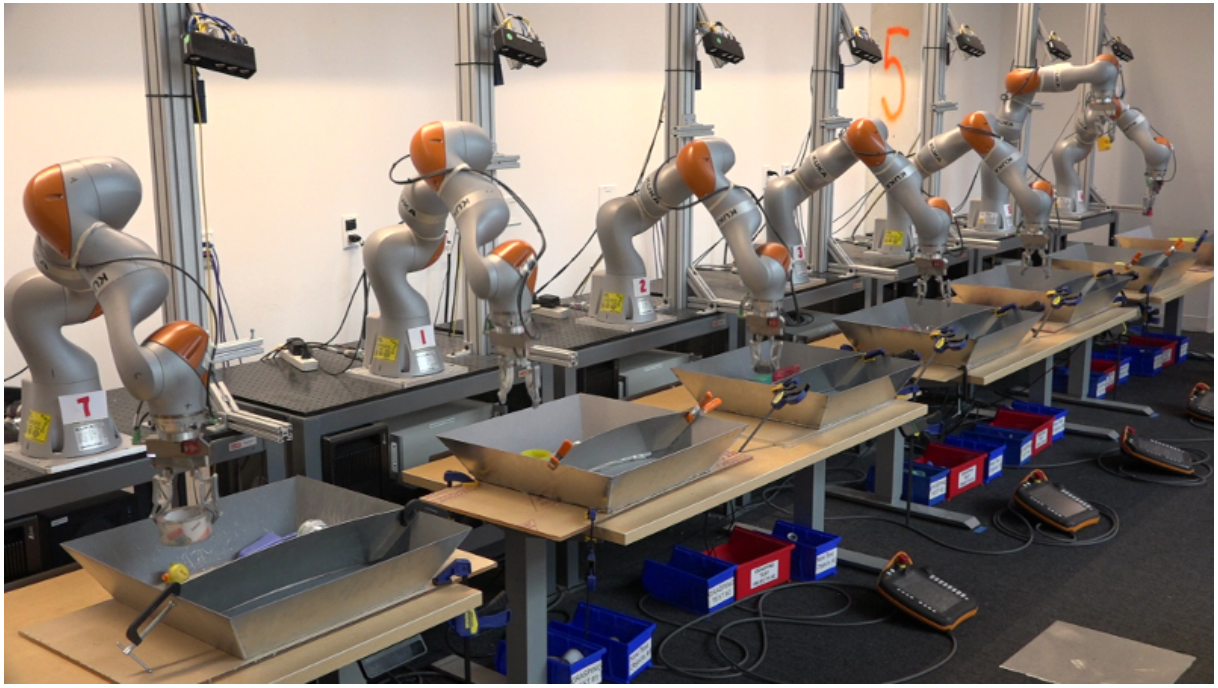}
    \caption{
       QT-opt \cite{qt_opt} collects grasping demonstration data with autonomous self-supervision.
    }
    \label{fig:qt_opt}
\end{figure}

One example of such a system is QT-opt \cite{qt_opt}, which collects self-supervised demonstration data for robot grasping in the real world. The system runs a reinforcement learning algorithm directly on the robot, receiving a binary reward for lifting an object successfully without any other reward shaping. Success is determined using a background subtraction test after the picked object is dropped. The setup for collecting data in QT-opt is shown in Figure \ref{fig:qt_opt}.

MT-opt \cite{mt_opt} is another demonstration collection system that continuously improves multi-task policies. Like QT-opt, it uses success detectors for self-supervised reward labeling, but in MT-opt, the success detectors are trained on data from all tasks and continuously updated to account for distribution shifts caused by factors such as varying lighting conditions and changing backgrounds in the real world. Additionally, MT-opt leverages the solutions to easier tasks to effectively bootstrap learning of more complex tasks, allowing the robot to train policies for harder tasks over time and collect better data for those tasks.

\subsection{Human Embodiment - Human Operator}

While robot demonstrations are naturally good for learning robotic tasks, there are a lot of works that try to let robots learn from human videos \cite{smith2019avid, yu2018one, nair2022r3m}. Therefore, in this section, we briefly discuss the data collection process of some representative datasets of human daily activities.

Ego4D \cite{ego4d} is a large-scale egocentric video dataset and benchmark suite that includes 3,670 hours of daily life activity videos captured by 931 unique camera wearers from 74 worldwide locations and 9 different countries. The dataset spans hundreds of scenarios, such as household, outdoor, workplace, leisure, and more. The key challenge of data collection is to recruit participants to record their daily activities. To overcome this challenge, the Ego4D project consists of 14 teams from universities and labs in 9 countries and 5 continents. Each team recruited participants to wear a camera for 1 to 10 hours at a time, and participants in 74 cities were recruited through word of mouth, ads, and postings on community bulletin boards.

RoboTube \cite{robotube} is a recently proposed dataset that provides a human video dataset and its digital twins for learning various robotic manipulation tasks. To collect the human video datasets, a portable video collection system with two RealSense D435 cameras was designed. The recording process streams two viewpoints: one is the first-person perspective from the camera mounted on the human head, and the other is the third-person perspective from the camera fixed on a tripod placed near the scene. These two streams are temporally synchronized, which enables the collection of high-quality video data for learning robotic manipulation tasks.
\section{Case Study: The Pipeline of Collecting and Utilizing Demonstrations in ManiSkill}
\label{sec:case_study}

\begin{figure*}[t!]
    \centering
    \includegraphics[width=0.95\linewidth]{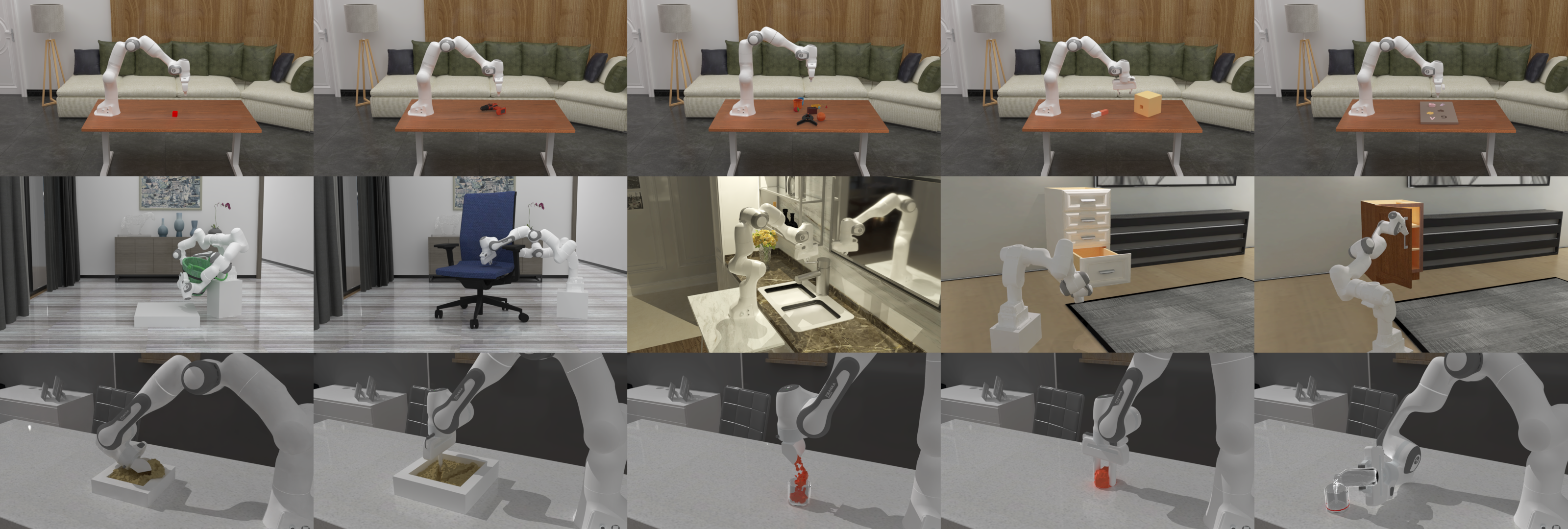}
    \caption{
       The tasks in ManiSkill2 \cite{maniskill2}.
    }
    \label{fig:ms2}
\end{figure*}

In this section, we present ManiSkill \cite{mu2021maniskill} (as well as its successor \cite{maniskill2}) as an example to demonstrate the entire pipeline of collecting and utilizing demonstrations. Henceforth, we refer to this series of works as ManiSkill for the sake of brevity.

ManiSkill is a benchmark for manipulation skills that can be generalized across a range of tasks. It comprises 20 manipulation task families with over 2000 object models and more than 4 million demonstration frames. These tasks include stationary/mobile-base, single/dual-arm, and rigid/soft-body manipulation, with 2D/3D-input data simulated by fully dynamic engines. It provides a unified interface and evaluation protocol to support a wide range of algorithms such as classic sense-plan-act, RL, and IL. It also accommodates various visual observations (point cloud, RGBD) and controllers (e.g., action type and parameterization). Fig \ref{fig:ms2} displays the visualization of tasks in ManiSkill2.

\subsection{Collecting Demonstrations}
\label{ms_demo_collection}

ManiSkill's approach of using different methods to generate demonstrations is motivated by the fact that the various tasks in the benchmark exhibit distinct characteristics, and thus, require tailored techniques. Specifically, task and motion planning (TAMP), model predictive control (MPC), and reinforcement learning (RL) are employed to generate demonstrations based on the task's difficulty and requirements.

TAMP is a suitable approach for many stationary manipulation tasks such as pick-and-place, as it does not require crafting rewards. However, it can face difficulties in dealing with underactuated systems like pushing chairs or moving buckets. In contrast, MPC is capable of searching for solutions to difficult tasks when given well-designed shaped rewards without the need for training or observations. However, designing a universal shaped reward for a variety of objects is non-trivial. On the other hand, RL requires additional training and hyperparameter tuning but is more scalable than MPC during inference.

For most rigid-body manipulation tasks and soft-body manipulation tasks, the demonstrations are generated through TAMP. However, for tasks involving manipulating articulated objects, the demonstration collection becomes more complex. Not only is manipulating articulated objects more challenging, but finding a universal solution for all object instances within a single task is also not straightforward.

For stationary manipulation tasks, such as turning a faucet, some faucet models can be solved by TAMP, while the remaining ones require solutions from MPC-CEM, utilizing designed dense rewards.

In the case of mobile manipulation tasks, such as pushing a swivel chair, TAMP is not a viable approach, and training a single RL agent to collect demonstrations for all objects is difficult. However, since training an agent to solve a specific object is feasible and well-studied, the demonstrations can be generated using a divide-and-conquer approach, wherein an RL agent is trained for each environment to generate successful demonstrations.

\subsection{Utilizing Demonstrations}

In ManiSkill, various demonstration-based baselines are evaluated to demonstrate the effectiveness of utilizing demonstrations.

To utilize demonstrations offline, ManiSkill benchmarks behavior cloning (BC), Batch-Constrained Q-Learning (BCQ) \cite{bcq}, and Twin-Delayed DDPG \cite{td3} with Behavior Cloning (TD3+BC) \cite{td3_bc}. The results show that when near-optimal demonstrations are available, BC achieves the strongest performance.

To augment online reinforcement learning with demonstrations, ManiSkill tests DAPG \cite{dapg}, which significantly outperforms the PPO \cite{ppo} baseline.

In addition, \cite{shen2022learning} investigates how to leverage large-scale demonstrations in ManiSkill by combining GAIL \cite{gail} and GASIL \cite{gasil} to utilize demonstrations in online reinforcement learning. The results show that combining offline imitation learning and online reinforcement learning yields much better object-level generalization in manipulation tasks.
\section{Conclusion \& Discussion}
\label{sec:discussion}

In conclusion, demonstrations have proven to be an effective way of improving the performance of agents in reinforcement learning and planning tasks. They offer valuable insights into the desired behavior of an agent, and can help accelerate the learning process. This survey has highlighted the various methods and approaches for utilizing demonstrations, especially in the context of embodied AI.

Despite the growing interest in demonstrations, there are still many challenges to overcome. One major issue is scaling up the demonstration collection process, which is currently hindered by limited scalability of teleoperation-based methods and quality control issues in learning-based autonomous pipelines. A promising solution could be the development of human-in-the-loop autonomous systems that can improve over time.

In addition to the collection of demonstrations, there is also a need to consider the types of demonstrations that are required. Questions such as the necessity of near-optimal demonstrations, the potential of learning from different embodiments, and the use of abstract demonstrations like natural language need to be addressed.

Another critical problem is the integration of offline and online learning. While some initial attempts have been made, there is still a need for solutions that can accommodate different forms and qualities of demonstrations in a variety of scenarios, particularly in partially observed and cross-domain contexts.

Overall, demonstrations have the potential to significantly enhance the performance of reinforcement learning and planning algorithms in real-world settings. However, there is still much to be explored and discovered in this area, and numerous opportunities exist for further research and development.


\bibliography{ref}

\begin{thebibliography}{100}
\providecommand{\url}[1]{#1}
\csname url@samestyle\endcsname
\providecommand{\newblock}{\relax}
\providecommand{\bibinfo}[2]{#2}
\providecommand{\BIBentrySTDinterwordspacing}{\spaceskip=0pt\relax}
\providecommand{\BIBentryALTinterwordstretchfactor}{4}
\providecommand{\BIBentryALTinterwordspacing}{\spaceskip=\fontdimen2\font plus
\BIBentryALTinterwordstretchfactor\fontdimen3\font minus
  \fontdimen4\font\relax}
\providecommand{\BIBforeignlanguage}[2]{{%
\expandafter\ifx\csname l@#1\endcsname\relax
\typeout{** WARNING: IEEEtran.bst: No hyphenation pattern has been}%
\typeout{** loaded for the language `#1'. Using the pattern for}%
\typeout{** the default language instead.}%
\else
\language=\csname l@#1\endcsname
\fi
#2}}
\providecommand{\BIBdecl}{\relax}
\BIBdecl

\bibitem{alphago}
D.~Silver, A.~Huang, C.~J. Maddison, A.~Guez, L.~Sifre, G.~Van Den~Driessche,
  J.~Schrittwieser, I.~Antonoglou, V.~Panneershelvam, M.~Lanctot \emph{et~al.},
  ``Mastering the game of go with deep neural networks and tree search,''
  \emph{nature}, vol. 529, no. 7587, pp. 484--489, 2016.

\bibitem{alphago_zero}
D.~Silver, J.~Schrittwieser, K.~Simonyan, I.~Antonoglou, A.~Huang, A.~Guez,
  T.~Hubert, L.~Baker, M.~Lai, A.~Bolton \emph{et~al.}, ``Mastering the game of
  go without human knowledge,'' \emph{nature}, vol. 550, no. 7676, pp.
  354--359, 2017.

\bibitem{dqn}
V.~Mnih, K.~Kavukcuoglu, D.~Silver, A.~A. Rusu, J.~Veness, M.~G. Bellemare,
  A.~Graves, M.~Riedmiller, A.~K. Fidjeland, G.~Ostrovski \emph{et~al.},
  ``Human-level control through deep reinforcement learning,'' \emph{nature},
  vol. 518, no. 7540, pp. 529--533, 2015.

\bibitem{openai_five}
C.~Berner, G.~Brockman, B.~Chan, V.~Cheung, P.~D{\k{e}}biak, C.~Dennison,
  D.~Farhi, Q.~Fischer, S.~Hashme, C.~Hesse \emph{et~al.}, ``Dota 2 with large
  scale deep reinforcement learning,'' \emph{arXiv preprint arXiv:1912.06680},
  2019.

\bibitem{hwangbo2019learning}
J.~Hwangbo, J.~Lee, A.~Dosovitskiy, D.~Bellicoso, V.~Tsounis, V.~Koltun, and
  M.~Hutter, ``Learning agile and dynamic motor skills for legged robots,''
  \emph{Science Robotics}, vol.~4, no.~26, p. eaau5872, 2019.

\bibitem{pets}
K.~Chua, R.~Calandra, R.~McAllister, and S.~Levine, ``Deep reinforcement
  learning in a handful of trials using probabilistic dynamics models,''
  \emph{Advances in neural information processing systems}, vol.~31, 2018.

\bibitem{planet}
D.~Hafner, T.~Lillicrap, I.~Fischer, R.~Villegas, D.~Ha, H.~Lee, and
  J.~Davidson, ``Learning latent dynamics for planning from pixels,'' in
  \emph{International conference on machine learning}.\hskip 1em plus 0.5em
  minus 0.4em\relax PMLR, 2019, pp. 2555--2565.

\bibitem{muzero}
J.~Schrittwieser, I.~Antonoglou, T.~Hubert, K.~Simonyan, L.~Sifre, S.~Schmitt,
  A.~Guez, E.~Lockhart, D.~Hassabis, T.~Graepel \emph{et~al.}, ``Mastering
  atari, go, chess and shogi by planning with a learned model,'' \emph{Nature},
  vol. 588, no. 7839, pp. 604--609, 2020.

\bibitem{efficient_zero}
W.~Ye, S.~Liu, T.~Kurutach, P.~Abbeel, and Y.~Gao, ``Mastering atari games with
  limited data,'' \emph{Advances in Neural Information Processing Systems},
  vol.~34, pp. 25\,476--25\,488, 2021.

\bibitem{td_mpc}
N.~Hansen, X.~Wang, and H.~Su, ``Temporal difference learning for model
  predictive control,'' \emph{arXiv preprint arXiv:2203.04955}, 2022.

\bibitem{syed2008apprenticeship}
U.~Syed, M.~Bowling, and R.~E. Schapire, ``Apprenticeship learning using linear
  programming,'' in \emph{Proceedings of the 25th international conference on
  Machine learning}, 2008, pp. 1032--1039.

\bibitem{sun2020scalability}
P.~Sun, H.~Kretzschmar, X.~Dotiwalla, A.~Chouard, V.~Patnaik, P.~Tsui, J.~Guo,
  Y.~Zhou, Y.~Chai, B.~Caine \emph{et~al.}, ``Scalability in perception for
  autonomous driving: Waymo open dataset,'' in \emph{Proceedings of the
  IEEE/CVF conference on computer vision and pattern recognition}, 2020, pp.
  2446--2454.

\bibitem{dasari2019robonet}
S.~Dasari, F.~Ebert, S.~Tian, S.~Nair, B.~Bucher, K.~Schmeckpeper, S.~Singh,
  S.~Levine, and C.~Finn, ``Robonet: Large-scale multi-robot learning,''
  \emph{arXiv preprint arXiv:1910.11215}, 2019.

\bibitem{saycan}
M.~Ahn, A.~Brohan, N.~Brown, Y.~Chebotar, O.~Cortes, B.~David, C.~Finn,
  K.~Gopalakrishnan, K.~Hausman, A.~Herzog \emph{et~al.}, ``Do as i can, not as
  i say: Grounding language in robotic affordances,'' \emph{arXiv preprint
  arXiv:2204.01691}, 2022.

\bibitem{rt1}
A.~Brohan, N.~Brown, J.~Carbajal, Y.~Chebotar, J.~Dabis, C.~Finn,
  K.~Gopalakrishnan, K.~Hausman, A.~Herzog, J.~Hsu \emph{et~al.}, ``Rt-1:
  Robotics transformer for real-world control at scale,'' \emph{arXiv preprint
  arXiv:2212.06817}, 2022.

\bibitem{mu2021maniskill}
T.~Mu, Z.~Ling, F.~Xiang, D.~C. Yang, X.~Li, S.~Tao, Z.~Huang, Z.~Jia, and
  H.~Su, ``Maniskill: Generalizable manipulation skill benchmark with
  large-scale demonstrations,'' in \emph{Thirty-fifth Conference on Neural
  Information Processing Systems Datasets and Benchmarks Track (Round 2)},
  2021.

\bibitem{rl_book}
R.~S. Sutton and A.~G. Barto, \emph{Reinforcement learning: An
  introduction}.\hskip 1em plus 0.5em minus 0.4em\relax MIT press, 2018.

\bibitem{bc}
D.~A. Pomerleau, ``Alvinn: An autonomous land vehicle in a neural network,''
  \emph{Advances in neural information processing systems}, vol.~1, 1988.

\bibitem{irl}
A.~Y. Ng, S.~J. Russell \emph{et~al.}, ``Algorithms for inverse reinforcement
  learning.'' in \emph{Icml}, vol.~1, 2000, p.~2.

\bibitem{gail}
J.~Ho and S.~Ermon, ``Generative adversarial imitation learning,''
  \emph{Advances in neural information processing systems}, vol.~29, 2016.

\bibitem{mpc_book}
E.~F. Camacho and C.~B. Alba, \emph{Model predictive control}.\hskip 1em plus
  0.5em minus 0.4em\relax Springer science \& business media, 2013.

\bibitem{mcts}
G.~Chaslot, S.~Bakkes, I.~Szita, and P.~Spronck, ``Monte-carlo tree search: A
  new framework for game ai,'' in \emph{Proceedings of the AAAI Conference on
  Artificial Intelligence and Interactive Digital Entertainment}, vol.~4,
  no.~1, 2008, pp. 216--217.

\bibitem{alpha_zero}
D.~Silver, T.~Hubert, J.~Schrittwieser, I.~Antonoglou, M.~Lai, A.~Guez,
  M.~Lanctot, L.~Sifre, D.~Kumaran, T.~Graepel \emph{et~al.}, ``Mastering chess
  and shogi by self-play with a general reinforcement learning algorithm,''
  \emph{arXiv preprint arXiv:1712.01815}, 2017.

\bibitem{bcq}
S.~Fujimoto, E.~Conti, M.~Ghavamzadeh, and J.~Pineau, ``Benchmarking batch deep
  reinforcement learning algorithms,'' \emph{arXiv preprint arXiv:1910.01708},
  2019.

\bibitem{bear}
A.~Kumar, J.~Fu, M.~Soh, G.~Tucker, and S.~Levine, ``Stabilizing off-policy
  q-learning via bootstrapping error reduction,'' \emph{Advances in Neural
  Information Processing Systems}, vol.~32, 2019.

\bibitem{brac}
Y.~Wu, G.~Tucker, and O.~Nachum, ``Behavior regularized offline reinforcement
  learning,'' \emph{arXiv preprint arXiv:1911.11361}, 2019.

\bibitem{rem}
R.~Agarwal, D.~Schuurmans, and M.~Norouzi, ``An optimistic perspective on
  offline reinforcement learning,'' in \emph{International Conference on
  Machine Learning}.\hskip 1em plus 0.5em minus 0.4em\relax PMLR, 2020, pp.
  104--114.

\bibitem{cql}
A.~Kumar, A.~Zhou, G.~Tucker, and S.~Levine, ``Conservative q-learning for
  offline reinforcement learning,'' \emph{Advances in Neural Information
  Processing Systems}, vol.~33, pp. 1179--1191, 2020.

\bibitem{distribution_shift}
S.~Ross, G.~Gordon, and D.~Bagnell, ``A reduction of imitation learning and
  structured prediction to no-regret online learning,'' in \emph{Proceedings of
  the fourteenth international conference on artificial intelligence and
  statistics}.\hskip 1em plus 0.5em minus 0.4em\relax JMLR Workshop and
  Conference Proceedings, 2011, pp. 627--635.

\bibitem{awac}
A.~Nair, A.~Gupta, M.~Dalal, and S.~Levine, ``Awac: Accelerating online
  reinforcement learning with offline datasets,'' \emph{arXiv preprint
  arXiv:2006.09359}, 2020.

\bibitem{aw_opt}
Y.~Lu, K.~Hausman, Y.~Chebotar, M.~Yan, E.~Jang, A.~Herzog, T.~Xiao, A.~Irpan,
  M.~Khansari, D.~Kalashnikov \emph{et~al.}, ``Aw-opt: Learning robotic skills
  with imitation andreinforcement at scale,'' in \emph{Conference on Robot
  Learning}.\hskip 1em plus 0.5em minus 0.4em\relax PMLR, 2022, pp. 1078--1088.

\bibitem{yinplanning}
Z.-H. Yin, W.~Ye, Q.~Chen, and Y.~Gao, ``Planning for sample efficient
  imitation learning,'' in \emph{Advances in Neural Information Processing
  Systems}.

\bibitem{niekum2013incremental}
S.~Niekum, S.~Chitta, A.~G. Barto, B.~Marthi, and S.~Osentoski, ``Incremental
  semantically grounded learning from demonstration.'' in \emph{Robotics:
  Science and Systems}, vol.~9.\hskip 1em plus 0.5em minus 0.4em\relax Berlin,
  Germany, 2013, pp. 10--15\,607.

\bibitem{fox2017multi}
R.~Fox, S.~Krishnan, I.~Stoica, and K.~Goldberg, ``Multi-level discovery of
  deep options,'' \emph{arXiv preprint arXiv:1703.08294}, 2017.

\bibitem{krishnan2017ddco}
S.~Krishnan, R.~Fox, I.~Stoica, and K.~Goldberg, ``Ddco: Discovery of deep
  continuous options for robot learning from demonstrations,'' in
  \emph{Conference on robot learning}.\hskip 1em plus 0.5em minus 0.4em\relax
  PMLR, 2017, pp. 418--437.

\bibitem{sharma2018directed}
A.~Sharma, M.~Sharma, N.~Rhinehart, and K.~M. Kitani, ``Directed-info gail:
  Learning hierarchical policies from unsegmented demonstrations using directed
  information,'' \emph{arXiv preprint arXiv:1810.01266}, 2018.

\bibitem{kipf2019compile}
T.~Kipf, Y.~Li, H.~Dai, V.~Zambaldi, A.~Sanchez-Gonzalez, E.~Grefenstette,
  P.~Kohli, and P.~Battaglia, ``Compile: Compositional imitation learning and
  execution,'' in \emph{International Conference on Machine Learning}.\hskip
  1em plus 0.5em minus 0.4em\relax PMLR, 2019, pp. 3418--3428.

\bibitem{bera2019podnet}
R.~Bera, V.~G. Goecks, G.~M. Gremillion, J.~Valasek, and N.~R. Waytowich,
  ``Podnet: A neural network for discovery of plannable options,'' \emph{arXiv
  preprint arXiv:1911.00171}, 2019.

\bibitem{pertsch2020accelerating}
K.~Pertsch, Y.~Lee, and J.~J. Lim, ``Accelerating reinforcement learning with
  learned skill priors,'' \emph{arXiv preprint arXiv:2010.11944}, 2020.

\bibitem{zhou2020plas}
W.~Zhou, S.~Bajracharya, and D.~Held, ``Plas: Latent action space for offline
  reinforcement learning,'' \emph{arXiv preprint arXiv:2011.07213}, 2020.

\bibitem{zhao2020augmenting}
Z.~Zhao, C.~Gan, J.~Wu, X.~Guo, and J.~B. Tenenbaum, ``Augmenting policy
  learning with routines discovered from a demonstration,'' \emph{arXiv
  preprint arXiv:2012.12469}, 2020.

\bibitem{lee2020learning}
S.-H. Lee and S.-W. Seo, ``Learning compound tasks without task-specific
  knowledge via imitation and self-supervised learning,'' in
  \emph{International Conference on Machine Learning}.\hskip 1em plus 0.5em
  minus 0.4em\relax PMLR, 2020, pp. 5747--5756.

\bibitem{ajay2020opal}
A.~Ajay, A.~Kumar, P.~Agrawal, S.~Levine, and O.~Nachum, ``Opal: Offline
  primitive discovery for accelerating offline reinforcement learning,''
  \emph{arXiv preprint arXiv:2010.13611}, 2020.

\bibitem{Shankar2020Discovering}
\BIBentryALTinterwordspacing
T.~Shankar, S.~Tulsiani, L.~Pinto, and A.~Gupta, ``Discovering motor programs
  by recomposing demonstrations,'' in \emph{International Conference on
  Learning Representations}, 2020. [Online]. Available:
  \url{https://openreview.net/forum?id=rkgHY0NYwr}
\BIBentrySTDinterwordspacing

\bibitem{shankar2020learning}
T.~Shankar and A.~Gupta, ``Learning robot skills with temporal variational
  inference,'' in \emph{International Conference on Machine Learning}.\hskip
  1em plus 0.5em minus 0.4em\relax PMLR, 2020, pp. 8624--8633.

\bibitem{lu2021learning}
Y.~Lu, Y.~Shen, S.~Zhou, A.~Courville, J.~B. Tenenbaum, and C.~Gan, ``Learning
  task decomposition with ordered memory policy network,'' \emph{arXiv preprint
  arXiv:2103.10972}, 2021.

\bibitem{zhu2021bottom}
Y.~Zhu, P.~Stone, and Y.~Zhu, ``Bottom-up skill discovery from unsegmented
  demonstrations for long-horizon robot manipulation,'' \emph{arXiv preprint
  arXiv:2109.13841}, 2021.

\bibitem{rao2021learning}
D.~Rao, F.~Sadeghi, L.~Hasenclever, M.~Wulfmeier, M.~Zambelli, G.~Vezzani,
  D.~Tirumala, Y.~Aytar, J.~Merel, N.~Heess \emph{et~al.}, ``Learning
  transferable motor skills with hierarchical latent mixture policies,''
  \emph{arXiv preprint arXiv:2112.05062}, 2021.

\bibitem{tanneberg2021skid}
D.~Tanneberg, K.~Ploeger, E.~Rueckert, and J.~Peters, ``Skid raw: Skill
  discovery from raw trajectories,'' \emph{IEEE Robotics and Automation
  Letters}, vol.~6, no.~3, pp. 4696--4703, 2021.

\bibitem{yang2021trail}
M.~Yang, S.~Levine, and O.~Nachum, ``Trail: Near-optimal imitation learning
  with suboptimal data,'' \emph{arXiv preprint arXiv:2110.14770}, 2021.

\bibitem{ghahramani2000variational}
Z.~Ghahramani and G.~E. Hinton, ``Variational learning for switching
  state-space models,'' \emph{Neural computation}, vol.~12, no.~4, pp.
  831--864, 2000.

\bibitem{blei2001topic}
D.~M. Blei and P.~J. Moreno, ``Topic segmentation with an aspect hidden markov
  model,'' in \emph{Proceedings of the 24th annual international ACM SIGIR
  conference on Research and development in information retrieval}, 2001, pp.
  343--348.

\bibitem{fox2011sticky}
E.~B. Fox, E.~B. Sudderth, M.~I. Jordan, and A.~S. Willsky, ``A sticky hdp-hmm
  with application to speaker diarization,'' \emph{The Annals of Applied
  Statistics}, pp. 1020--1056, 2011.

\bibitem{linderman2017bayesian}
S.~Linderman, M.~Johnson, A.~Miller, R.~Adams, D.~Blei, and L.~Paninski,
  ``Bayesian learning and inference in recurrent switching linear dynamical
  systems,'' in \emph{Artificial Intelligence and Statistics}.\hskip 1em plus
  0.5em minus 0.4em\relax PMLR, 2017, pp. 914--922.

\bibitem{love}
Y.~Jiang, E.~Z. Liu, B.~Eysenbach, Z.~Kolter, and C.~Finn, ``Learning options
  via compression,'' \emph{arXiv preprint arXiv:2212.04590}, 2022.

\bibitem{franka_kitchen}
A.~Gupta, V.~Kumar, C.~Lynch, S.~Levine, and K.~Hausman, ``Relay policy
  learning: Solving long-horizon tasks via imitation and reinforcement
  learning,'' \emph{arXiv preprint arXiv:1910.11956}, 2019.

\bibitem{andrychowicz2017hindsight}
M.~Andrychowicz, F.~Wolski, A.~Ray, J.~Schneider, R.~Fong, P.~Welinder,
  B.~McGrew, J.~Tobin, O.~Pieter~Abbeel, and W.~Zaremba, ``Hindsight experience
  replay,'' \emph{Advances in neural information processing systems}, vol.~30,
  2017.

\bibitem{singh2020parrot}
A.~Singh, H.~Liu, G.~Zhou, A.~Yu, N.~Rhinehart, and S.~Levine, ``Parrot:
  Data-driven behavioral priors for reinforcement learning,'' \emph{arXiv
  preprint arXiv:2011.10024}, 2020.

\bibitem{me_trpo}
T.~Kurutach, I.~Clavera, Y.~Duan, A.~Tamar, and P.~Abbeel, ``Model-ensemble
  trust-region policy optimization,'' \emph{arXiv preprint arXiv:1802.10592},
  2018.

\bibitem{dreamer}
D.~Hafner, T.~Lillicrap, J.~Ba, and M.~Norouzi, ``Dream to control: Learning
  behaviors by latent imagination,'' \emph{arXiv preprint arXiv:1912.01603},
  2019.

\bibitem{mbrl_atari}
L.~Kaiser, M.~Babaeizadeh, P.~Milos, B.~Osinski, R.~H. Campbell, K.~Czechowski,
  D.~Erhan, C.~Finn, P.~Kozakowski, S.~Levine \emph{et~al.}, ``Model-based
  reinforcement learning for atari,'' \emph{arXiv preprint arXiv:1903.00374},
  2019.

\bibitem{ross2012agnostic}
S.~Ross and J.~A. Bagnell, ``Agnostic system identification for model-based
  reinforcement learning,'' \emph{arXiv preprint arXiv:1203.1007}, 2012.

\bibitem{kidambi2020morel}
R.~Kidambi, A.~Rajeswaran, P.~Netrapalli, and T.~Joachims, ``Morel: Model-based
  offline reinforcement learning,'' \emph{Advances in neural information
  processing systems}, vol.~33, pp. 21\,810--21\,823, 2020.

\bibitem{yu2020mopo}
T.~Yu, G.~Thomas, L.~Yu, S.~Ermon, J.~Y. Zou, S.~Levine, C.~Finn, and T.~Ma,
  ``Mopo: Model-based offline policy optimization,'' \emph{Advances in Neural
  Information Processing Systems}, vol.~33, pp. 14\,129--14\,142, 2020.

\bibitem{xie2018few}
A.~Xie, A.~Singh, S.~Levine, and C.~Finn, ``Few-shot goal inference for
  visuomotor learning and planning,'' in \emph{Conference on Robot
  Learning}.\hskip 1em plus 0.5em minus 0.4em\relax PMLR, 2018, pp. 40--52.

\bibitem{vecerik2019practical}
M.~Vecerik, O.~Sushkov, D.~Barker, T.~Roth{\"o}rl, T.~Hester, and J.~Scholz,
  ``A practical approach to insertion with variable socket position using deep
  reinforcement learning,'' in \emph{2019 international conference on robotics
  and automation (ICRA)}.\hskip 1em plus 0.5em minus 0.4em\relax IEEE, 2019,
  pp. 754--760.

\bibitem{singh2019end}
A.~Singh, L.~Yang, K.~Hartikainen, C.~Finn, and S.~Levine, ``End-to-end robotic
  reinforcement learning without reward engineering,'' \emph{arXiv preprint
  arXiv:1904.07854}, 2019.

\bibitem{zolna2020offline}
K.~Zolna, A.~Novikov, K.~Konyushkova, C.~Gulcehre, Z.~Wang, Y.~Aytar, M.~Denil,
  N.~de~Freitas, and S.~Reed, ``Offline learning from demonstrations and
  unlabeled experience,'' \emph{arXiv preprint arXiv:2011.13885}, 2020.

\bibitem{simclr}
T.~Chen, S.~Kornblith, M.~Norouzi, and G.~Hinton, ``A simple framework for
  contrastive learning of visual representations,'' in \emph{International
  conference on machine learning}.\hskip 1em plus 0.5em minus 0.4em\relax PMLR,
  2020, pp. 1597--1607.

\bibitem{moco}
K.~He, H.~Fan, Y.~Wu, S.~Xie, and R.~Girshick, ``Momentum contrast for
  unsupervised visual representation learning,'' in \emph{Proceedings of the
  IEEE/CVF conference on computer vision and pattern recognition}, 2020, pp.
  9729--9738.

\bibitem{doersch2015unsupervised}
C.~Doersch, A.~Gupta, and A.~A. Efros, ``Unsupervised visual representation
  learning by context prediction,'' in \emph{Proceedings of the IEEE
  international conference on computer vision}, 2015, pp. 1422--1430.

\bibitem{oord2018representation}
A.~v.~d. Oord, Y.~Li, and O.~Vinyals, ``Representation learning with
  contrastive predictive coding,'' \emph{arXiv preprint arXiv:1807.03748},
  2018.

\bibitem{devlin2018bert}
J.~Devlin, M.-W. Chang, K.~Lee, and K.~Toutanova, ``Bert: Pre-training of deep
  bidirectional transformers for language understanding,'' \emph{arXiv preprint
  arXiv:1810.04805}, 2018.

\bibitem{brown2020language}
T.~Brown, B.~Mann, N.~Ryder, M.~Subbiah, J.~D. Kaplan, P.~Dhariwal,
  A.~Neelakantan, P.~Shyam, G.~Sastry, A.~Askell \emph{et~al.}, ``Language
  models are few-shot learners,'' \emph{Advances in neural information
  processing systems}, vol.~33, pp. 1877--1901, 2020.

\bibitem{chowdhery2022palm}
A.~Chowdhery, S.~Narang, J.~Devlin, M.~Bosma, G.~Mishra, A.~Roberts, P.~Barham,
  H.~W. Chung, C.~Sutton, S.~Gehrmann \emph{et~al.}, ``Palm: Scaling language
  modeling with pathways,'' \emph{arXiv preprint arXiv:2204.02311}, 2022.

\bibitem{pari2021surprising}
J.~Pari, N.~M. Shafiullah, S.~P. Arunachalam, and L.~Pinto, ``The surprising
  effectiveness of representation learning for visual imitation,'' \emph{arXiv
  preprint arXiv:2112.01511}, 2021.

\bibitem{yang2021representation}
M.~Yang and O.~Nachum, ``Representation matters: offline pretraining for
  sequential decision making,'' in \emph{International Conference on Machine
  Learning}.\hskip 1em plus 0.5em minus 0.4em\relax PMLR, 2021, pp.
  11\,784--11\,794.

\bibitem{zhan2022learning}
A.~Zhan, R.~Zhao, L.~Pinto, P.~Abbeel, and M.~Laskin, ``Learning visual robotic
  control efficiently with contrastive pre-training and data augmentation,'' in
  \emph{2022 IEEE/RSJ International Conference on Intelligent Robots and
  Systems (IROS)}.\hskip 1em plus 0.5em minus 0.4em\relax IEEE, 2022, pp.
  4040--4047.

\bibitem{shah2021rrl}
R.~Shah and V.~Kumar, ``Rrl: Resnet as representation for reinforcement
  learning,'' \emph{arXiv preprint arXiv:2107.03380}, 2021.

\bibitem{nair2022r3m}
S.~Nair, A.~Rajeswaran, V.~Kumar, C.~Finn, and A.~Gupta, ``R3m: A universal
  visual representation for robot manipulation,'' \emph{arXiv preprint
  arXiv:2203.12601}, 2022.

\bibitem{wang2022vrl3}
C.~Wang, X.~Luo, K.~Ross, and D.~Li, ``Vrl3: A data-driven framework for visual
  deep reinforcement learning,'' \emph{arXiv preprint arXiv:2202.10324}, 2022.

\bibitem{mvp2}
I.~Radosavovic, T.~Xiao, S.~James, P.~Abbeel, J.~Malik, and T.~Darrell,
  ``Real-world robot learning with masked visual pre-training,'' \emph{arXiv
  preprint arXiv:2210.03109}, 2022.

\bibitem{one_shot_il}
Y.~Duan, M.~Andrychowicz, B.~Stadie, O.~Jonathan~Ho, J.~Schneider,
  I.~Sutskever, P.~Abbeel, and W.~Zaremba, ``One-shot imitation learning,''
  \emph{Advances in neural information processing systems}, vol.~30, 2017.

\bibitem{finn2017one}
C.~Finn, T.~Yu, T.~Zhang, P.~Abbeel, and S.~Levine, ``One-shot visual imitation
  learning via meta-learning,'' in \emph{Conference on robot learning}.\hskip
  1em plus 0.5em minus 0.4em\relax PMLR, 2017, pp. 357--368.

\bibitem{pathak2018zero}
D.~Pathak, P.~Mahmoudieh, G.~Luo, P.~Agrawal, D.~Chen, Y.~Shentu, E.~Shelhamer,
  J.~Malik, A.~A. Efros, and T.~Darrell, ``Zero-shot visual imitation,'' in
  \emph{Proceedings of the IEEE conference on computer vision and pattern
  recognition workshops}, 2018, pp. 2050--2053.

\bibitem{yu2018one}
T.~Yu, C.~Finn, A.~Xie, S.~Dasari, T.~Zhang, P.~Abbeel, and S.~Levine,
  ``One-shot imitation from observing humans via domain-adaptive
  meta-learning,'' \emph{arXiv preprint arXiv:1802.01557}, 2018.

\bibitem{lynch2020language}
C.~Lynch and P.~Sermanet, ``Language conditioned imitation learning over
  unstructured data,'' \emph{arXiv preprint arXiv:2005.07648}, 2020.

\bibitem{wei2022chain}
J.~Wei, X.~Wang, D.~Schuurmans, M.~Bosma, E.~Chi, Q.~Le, and D.~Zhou, ``Chain
  of thought prompting elicits reasoning in large language models,''
  \emph{arXiv preprint arXiv:2201.11903}, 2022.

\bibitem{gato}
S.~Reed, K.~Zolna, E.~Parisotto, S.~G. Colmenarejo, A.~Novikov, G.~Barth-Maron,
  M.~Gimenez, Y.~Sulsky, J.~Kay, J.~T. Springenberg \emph{et~al.}, ``A
  generalist agent,'' \emph{arXiv preprint arXiv:2205.06175}, 2022.

\bibitem{laskin2022context}
M.~Laskin, L.~Wang, J.~Oh, E.~Parisotto, S.~Spencer, R.~Steigerwald,
  D.~Strouse, S.~Hansen, A.~Filos, E.~Brooks \emph{et~al.}, ``In-context
  reinforcement learning with algorithm distillation,'' \emph{arXiv preprint
  arXiv:2210.14215}, 2022.

\bibitem{rainbow}
M.~Hessel, J.~Modayil, H.~Van~Hasselt, T.~Schaul, G.~Ostrovski, W.~Dabney,
  D.~Horgan, B.~Piot, M.~Azar, and D.~Silver, ``Rainbow: Combining improvements
  in deep reinforcement learning,'' in \emph{Thirty-second AAAI conference on
  artificial intelligence}, 2018.

\bibitem{ddpg}
T.~P. Lillicrap, J.~J. Hunt, A.~Pritzel, N.~Heess, T.~Erez, Y.~Tassa,
  D.~Silver, and D.~Wierstra, ``Continuous control with deep reinforcement
  learning,'' \emph{arXiv preprint arXiv:1509.02971}, 2015.

\bibitem{td3}
S.~Fujimoto, H.~Hoof, and D.~Meger, ``Addressing function approximation error
  in actor-critic methods,'' in \emph{International conference on machine
  learning}.\hskip 1em plus 0.5em minus 0.4em\relax PMLR, 2018, pp. 1587--1596.

\bibitem{sac}
T.~Haarnoja, A.~Zhou, P.~Abbeel, and S.~Levine, ``Soft actor-critic: Off-policy
  maximum entropy deep reinforcement learning with a stochastic actor,'' in
  \emph{International conference on machine learning}.\hskip 1em plus 0.5em
  minus 0.4em\relax PMLR, 2018, pp. 1861--1870.

\bibitem{dqnfd}
T.~Hester, M.~Vecerik, O.~Pietquin, M.~Lanctot, T.~Schaul, B.~Piot, D.~Horgan,
  J.~Quan, A.~Sendonaris, I.~Osband \emph{et~al.}, ``Deep q-learning from
  demonstrations,'' in \emph{Proceedings of the AAAI Conference on Artificial
  Intelligence}, vol.~32, no.~1, 2018.

\bibitem{r2d2}
S.~Kapturowski, G.~Ostrovski, J.~Quan, R.~Munos, and W.~Dabney, ``Recurrent
  experience replay in distributed reinforcement learning,'' in
  \emph{International conference on learning representations}, 2018.

\bibitem{r2d3}
T.~L. Paine, C.~Gulcehre, B.~Shahriari, M.~Denil, M.~Hoffman, H.~Soyer,
  R.~Tanburn, S.~Kapturowski, N.~Rabinowitz, D.~Williams \emph{et~al.},
  ``Making efficient use of demonstrations to solve hard exploration
  problems,'' \emph{arXiv preprint arXiv:1909.01387}, 2019.

\bibitem{ddpgfd}
M.~Vecerik, T.~Hester, J.~Scholz, F.~Wang, O.~Pietquin, B.~Piot, N.~Heess,
  T.~Roth{\"o}rl, T.~Lampe, and M.~Riedmiller, ``Leveraging demonstrations for
  deep reinforcement learning on robotics problems with sparse rewards,''
  \emph{arXiv preprint arXiv:1707.08817}, 2017.

\bibitem{nair2018overcoming}
A.~Nair, B.~McGrew, M.~Andrychowicz, W.~Zaremba, and P.~Abbeel, ``Overcoming
  exploration in reinforcement learning with demonstrations,'' in \emph{2018
  IEEE international conference on robotics and automation (ICRA)}.\hskip 1em
  plus 0.5em minus 0.4em\relax IEEE, 2018, pp. 6292--6299.

\bibitem{pofd}
B.~Kang, Z.~Jie, and J.~Feng, ``Policy optimization with demonstrations,'' in
  \emph{International conference on machine learning}.\hskip 1em plus 0.5em
  minus 0.4em\relax PMLR, 2018, pp. 2469--2478.

\bibitem{max_entropy_irl}
B.~D. Ziebart, A.~L. Maas, J.~A. Bagnell, A.~K. Dey \emph{et~al.}, ``Maximum
  entropy inverse reinforcement learning.'' in \emph{Aaai}, vol.~8.\hskip 1em
  plus 0.5em minus 0.4em\relax Chicago, IL, USA, 2008, pp. 1433--1438.

\bibitem{sqil}
S.~Reddy, A.~D. Dragan, and S.~Levine, ``Sqil: Imitation learning via
  reinforcement learning with sparse rewards,'' \emph{arXiv preprint
  arXiv:1905.11108}, 2019.

\bibitem{duan2016benchmarking}
Y.~Duan, X.~Chen, R.~Houthooft, J.~Schulman, and P.~Abbeel, ``Benchmarking deep
  reinforcement learning for continuous control,'' in \emph{International
  conference on machine learning}.\hskip 1em plus 0.5em minus 0.4em\relax PMLR,
  2016, pp. 1329--1338.

\bibitem{drl_matters}
P.~Henderson, R.~Islam, P.~Bachman, J.~Pineau, D.~Precup, and D.~Meger, ``Deep
  reinforcement learning that matters,'' in \emph{Proceedings of the AAAI
  conference on artificial intelligence}, vol.~32, no.~1, 2018.

\bibitem{trpo}
J.~Schulman, S.~Levine, P.~Abbeel, M.~Jordan, and P.~Moritz, ``Trust region
  policy optimization,'' in \emph{International conference on machine
  learning}.\hskip 1em plus 0.5em minus 0.4em\relax PMLR, 2015, pp. 1889--1897.

\bibitem{ppo}
J.~Schulman, F.~Wolski, P.~Dhariwal, A.~Radford, and O.~Klimov, ``Proximal
  policy optimization algorithms,'' \emph{arXiv preprint arXiv:1707.06347},
  2017.

\bibitem{dapg}
A.~Rajeswaran, V.~Kumar, A.~Gupta, G.~Vezzani, J.~Schulman, E.~Todorov, and
  S.~Levine, ``Learning complex dexterous manipulation with deep reinforcement
  learning and demonstrations,'' \emph{arXiv preprint arXiv:1709.10087}, 2017.

\bibitem{aytar2018playing}
Y.~Aytar, T.~Pfaff, D.~Budden, T.~Paine, Z.~Wang, and N.~De~Freitas, ``Playing
  hard exploration games by watching youtube,'' \emph{Advances in neural
  information processing systems}, vol.~31, 2018.

\bibitem{smith2019avid}
L.~Smith, N.~Dhawan, M.~Zhang, P.~Abbeel, and S.~Levine, ``Avid: Learning
  multi-stage tasks via pixel-level translation of human videos,'' \emph{arXiv
  preprint arXiv:1912.04443}, 2019.

\bibitem{brys2015reinforcement}
T.~Brys, A.~Harutyunyan, H.~B. Suay, S.~Chernova, M.~E. Taylor, and
  A.~Now{\'e}, ``Reinforcement learning from demonstration through shaping,''
  in \emph{Twenty-fourth international joint conference on artificial
  intelligence}, 2015.

\bibitem{ng1999policy}
A.~Y. Ng, D.~Harada, and S.~Russell, ``Policy invariance under reward
  transformations: Theory and application to reward shaping,'' in \emph{Icml},
  vol.~99, 1999, pp. 278--287.

\bibitem{wu2021learning}
Z.~Wu, W.~Lian, V.~Unhelkar, M.~Tomizuka, and S.~Schaal, ``Learning dense
  rewards for contact-rich manipulation tasks,'' in \emph{2021 IEEE
  International Conference on Robotics and Automation (ICRA)}.\hskip 1em plus
  0.5em minus 0.4em\relax IEEE, 2021, pp. 6214--6221.

\bibitem{graph_irl}
S.~Kumar, J.~Zamora, N.~Hansen, R.~Jangir, and X.~Wang, ``Graph inverse
  reinforcement learning from diverse videos,'' \emph{arXiv preprint
  arXiv:2207.14299}, 2022.

\bibitem{peng2018deepmimic}
X.~B. Peng, P.~Abbeel, S.~Levine, and M.~Van~de Panne, ``Deepmimic:
  Example-guided deep reinforcement learning of physics-based character
  skills,'' \emph{ACM Transactions On Graphics (TOG)}, vol.~37, no.~4, pp.
  1--14, 2018.

\bibitem{peng2020learning}
X.~B. Peng, E.~Coumans, T.~Zhang, T.-W. Lee, J.~Tan, and S.~Levine, ``Learning
  agile robotic locomotion skills by imitating animals,'' \emph{arXiv preprint
  arXiv:2004.00784}, 2020.

\bibitem{peng2022ase}
X.~B. Peng, Y.~Guo, L.~Halper, S.~Levine, and S.~Fidler, ``Ase: Large-scale
  reusable adversarial skill embeddings for physically simulated characters,''
  \emph{arXiv preprint arXiv:2205.01906}, 2022.

\bibitem{irl_survey}
S.~Arora and P.~Doshi, ``A survey of inverse reinforcement learning:
  Challenges, methods and progress,'' \emph{Artificial Intelligence}, vol. 297,
  p. 103500, 2021.

\bibitem{algo_irl}
A.~Y. Ng, S.~Russell \emph{et~al.}, ``Algorithms for inverse reinforcement
  learning.'' in \emph{Icml}, vol.~1, 2000, p.~2.

\bibitem{abbeel2004apprenticeship}
P.~Abbeel and A.~Y. Ng, ``Apprenticeship learning via inverse reinforcement
  learning,'' in \emph{Proceedings of the twenty-first international conference
  on Machine learning}, 2004, p.~1.

\bibitem{mdp_book}
M.~L. Puterman, \emph{Markov decision processes: discrete stochastic dynamic
  programming}.\hskip 1em plus 0.5em minus 0.4em\relax John Wiley \& Sons,
  2014.

\bibitem{dac}
I.~Kostrikov, K.~K. Agrawal, D.~Dwibedi, S.~Levine, and J.~Tompson,
  ``Discriminator-actor-critic: Addressing sample inefficiency and reward bias
  in adversarial imitation learning,'' \emph{arXiv preprint arXiv:1809.02925},
  2018.

\bibitem{airl}
J.~Fu, K.~Luo, and S.~Levine, ``Learning robust rewards with adversarial
  inverse reinforcement learning,'' \emph{arXiv preprint arXiv:1710.11248},
  2017.

\bibitem{ghasemipour2020divergence}
S.~K.~S. Ghasemipour, R.~Zemel, and S.~Gu, ``A divergence minimization
  perspective on imitation learning methods,'' in \emph{Conference on Robot
  Learning}.\hskip 1em plus 0.5em minus 0.4em\relax PMLR, 2020, pp. 1259--1277.

\bibitem{gan}
I.~Goodfellow, J.~Pouget-Abadie, M.~Mirza, B.~Xu, D.~Warde-Farley, S.~Ozair,
  A.~Courville, and Y.~Bengio, ``Generative adversarial networks,''
  \emph{Communications of the ACM}, vol.~63, no.~11, pp. 139--144, 2020.

\bibitem{pwil}
R.~Dadashi, L.~Hussenot, M.~Geist, and O.~Pietquin, ``Primal wasserstein
  imitation learning,'' \emph{arXiv preprint arXiv:2006.04678}, 2020.

\bibitem{hosu2016playing}
I.-A. Hosu and T.~Rebedea, ``Playing atari games with deep reinforcement
  learning and human checkpoint replay,'' \emph{arXiv preprint
  arXiv:1607.05077}, 2016.

\bibitem{salimans2018learning}
T.~Salimans and R.~Chen, ``Learning montezuma's revenge from a single
  demonstration,'' \emph{arXiv preprint arXiv:1812.03381}, 2018.

\bibitem{zhu2018reinforcement}
Y.~Zhu, Z.~Wang, J.~Merel, A.~Rusu, T.~Erez, S.~Cabi, S.~Tunyasuvunakool,
  J.~Kram{\'a}r, R.~Hadsell, N.~de~Freitas \emph{et~al.}, ``Reinforcement and
  imitation learning for diverse visuomotor skills,'' \emph{arXiv preprint
  arXiv:1802.09564}, 2018.

\bibitem{rfuniverse}
H.~Fu, W.~Xu, H.~Xue, H.~Yang, R.~Ye, Y.~Huang, Z.~Xue, Y.~Wang, and C.~Lu,
  ``Rfuniverse: A physics-based action-centric interactive environment for
  everyday household tasks,'' \emph{arXiv preprint arXiv:2202.00199}, 2022.

\bibitem{zhu2020robosuite}
Y.~Zhu, J.~Wong, A.~Mandlekar, and R.~Mart{\'\i}n-Mart{\'\i}n, ``robosuite: A
  modular simulation framework and benchmark for robot learning,'' \emph{arXiv
  preprint arXiv:2009.12293}, 2020.

\bibitem{igibson}
B.~Shen, F.~Xia, C.~Li, R.~Mart{\'\i}n-Mart{\'\i}n, L.~Fan, G.~Wang,
  C.~P{\'e}rez-D’Arpino, S.~Buch, S.~Srivastava, L.~Tchapmi \emph{et~al.},
  ``igibson 1.0: a simulation environment for interactive tasks in large
  realistic scenes,'' in \emph{2021 IEEE/RSJ International Conference on
  Intelligent Robots and Systems (IROS)}.\hskip 1em plus 0.5em minus
  0.4em\relax IEEE, 2021, pp. 7520--7527.

\bibitem{roboturk}
A.~Mandlekar, Y.~Zhu, A.~Garg, J.~Booher, M.~Spero, A.~Tung, J.~Gao, J.~Emmons,
  A.~Gupta, E.~Orbay \emph{et~al.}, ``Roboturk: A crowdsourcing platform for
  robotic skill learning through imitation,'' in \emph{Conference on Robot
  Learning}.\hskip 1em plus 0.5em minus 0.4em\relax PMLR, 2018, pp. 879--893.

\bibitem{mujoco_vr}
V.~Kumar and E.~Todorov, ``Mujoco haptix: A virtual reality system for hand
  manipulation,'' in \emph{2015 IEEE-RAS 15th International Conference on
  Humanoid Robots (Humanoids)}.\hskip 1em plus 0.5em minus 0.4em\relax IEEE,
  2015, pp. 657--663.

\bibitem{igibson2}
C.~Li, F.~Xia, R.~Mart{\'\i}n-Mart{\'\i}n, M.~Lingelbach, S.~Srivastava,
  B.~Shen, K.~Vainio, C.~Gokmen, G.~Dharan, T.~Jain \emph{et~al.}, ``igibson
  2.0: Object-centric simulation for robot learning of everyday household
  tasks,'' \emph{arXiv preprint arXiv:2108.03272}, 2021.

\bibitem{openvr}
W.~Wang, Y.~Suga, H.~Iwata, and S.~Sugano, ``Openvr: A software tool
  contributes to research of robotics,'' in \emph{2011 IEEE/SICE International
  Symposium on System Integration (SII)}.\hskip 1em plus 0.5em minus
  0.4em\relax IEEE, 2011, pp. 1043--1048.

\bibitem{puig2020watch}
X.~Puig, T.~Shu, S.~Li, Z.~Wang, Y.-H. Liao, J.~B. Tenenbaum, S.~Fidler, and
  A.~Torralba, ``Watch-and-help: A challenge for social perception and human-ai
  collaboration,'' \emph{arXiv preprint arXiv:2010.09890}, 2020.

\bibitem{shridhar2020alfred}
M.~Shridhar, J.~Thomason, D.~Gordon, Y.~Bisk, W.~Han, R.~Mottaghi,
  L.~Zettlemoyer, and D.~Fox, ``Alfred: A benchmark for interpreting grounded
  instructions for everyday tasks,'' in \emph{Proceedings of the IEEE/CVF
  conference on computer vision and pattern recognition}, 2020, pp.
  10\,740--10\,749.

\bibitem{hoffmann2001ff}
J.~Hoffmann and B.~Nebel, ``The ff planning system: Fast plan generation
  through heuristic search,'' \emph{Journal of Artificial Intelligence
  Research}, vol.~14, pp. 253--302, 2001.

\bibitem{kuffner2000rrt}
J.~J. Kuffner and S.~M. LaValle, ``Rrt-connect: An efficient approach to
  single-query path planning,'' in \emph{Proceedings 2000 ICRA. Millennium
  Conference. IEEE International Conference on Robotics and Automation.
  Symposia Proceedings (Cat. No. 00CH37065)}, vol.~2.\hskip 1em plus 0.5em
  minus 0.4em\relax IEEE, 2000, pp. 995--1001.

\bibitem{prm}
R.~Bohlin and L.~E. Kavraki, ``Path planning using lazy prm,'' in
  \emph{Proceedings 2000 ICRA. Millennium conference. IEEE international
  conference on robotics and automation. Symposia proceedings (Cat. No.
  00CH37065)}, vol.~1.\hskip 1em plus 0.5em minus 0.4em\relax IEEE, 2000, pp.
  521--528.

\bibitem{rlbench}
S.~James, Z.~Ma, D.~R. Arrojo, and A.~J. Davison, ``Rlbench: The robot learning
  benchmark \& learning environment,'' \emph{IEEE Robotics and Automation
  Letters}, vol.~5, no.~2, pp. 3019--3026, 2020.

\bibitem{d4rl}
J.~Fu, A.~Kumar, O.~Nachum, G.~Tucker, and S.~Levine, ``D4rl: Datasets for deep
  data-driven reinforcement learning,'' \emph{arXiv preprint arXiv:2004.07219},
  2020.

\bibitem{qt_opt}
D.~Kalashnikov, A.~Irpan, P.~Pastor, J.~Ibarz, A.~Herzog, E.~Jang, D.~Quillen,
  E.~Holly, M.~Kalakrishnan, V.~Vanhoucke \emph{et~al.}, ``Scalable deep
  reinforcement learning for vision-based robotic manipulation,'' in
  \emph{Conference on Robot Learning}.\hskip 1em plus 0.5em minus 0.4em\relax
  PMLR, 2018, pp. 651--673.

\bibitem{mt_opt}
D.~Kalashnikov, J.~Varley, Y.~Chebotar, B.~Swanson, R.~Jonschkowski, C.~Finn,
  S.~Levine, and K.~Hausman, ``Mt-opt: Continuous multi-task robotic
  reinforcement learning at scale,'' \emph{arXiv preprint arXiv:2104.08212},
  2021.

\bibitem{ego4d}
K.~Grauman, A.~Westbury, E.~Byrne, Z.~Chavis, A.~Furnari, R.~Girdhar,
  J.~Hamburger, H.~Jiang, M.~Liu, X.~Liu \emph{et~al.}, ``Ego4d: Around the
  world in 3,000 hours of egocentric video,'' in \emph{Proceedings of the
  IEEE/CVF Conference on Computer Vision and Pattern Recognition}, 2022, pp.
  18\,995--19\,012.

\bibitem{robotube}
\BIBentryALTinterwordspacing
haoyu Xiong, H.~Fu, J.~Zhang, C.~Bao, Q.~Zhang, Y.~Huang, W.~Xu, A.~Garg, and
  C.~Lu, ``Robotube: Learning household manipulation from human videos with
  simulated twin environments,'' in \emph{6th Annual Conference on Robot
  Learning}, 2022. [Online]. Available:
  \url{https://openreview.net/forum?id=VD0nXUG5Qk}
\BIBentrySTDinterwordspacing

\bibitem{maniskill2}
J.~Gu, F.~Xiang, X.~Li, Z.~Ling, X.~Liu, T.~Mu, Y.~Tang, S.~Tao, X.~Wei,
  Y.~Yao, X.~Yuan, P.~Xie, Z.~Huang, R.~Chen, and H.~Su, ``Maniskill2: A
  unified benchmark for generalizable manipulation skills,'' in
  \emph{International Conference on Learning Representations}, 2023.

\bibitem{td3_bc}
S.~Fujimoto and S.~S. Gu, ``A minimalist approach to offline reinforcement
  learning,'' \emph{Advances in neural information processing systems},
  vol.~34, pp. 20\,132--20\,145, 2021.

\bibitem{shen2022learning}
H.~Shen, W.~Wan, and H.~Wang, ``Learning category-level generalizable object
  manipulation policy via generative adversarial self-imitation learning from
  demonstrations,'' \emph{arXiv preprint arXiv:2203.02107}, 2022.

\bibitem{gasil}
Y.~Guo, J.~Oh, S.~Singh, and H.~Lee, ``Generative adversarial self-imitation
  learning,'' \emph{arXiv preprint arXiv:1812.00950}, 2018.

\end{thebibliography}
\bibliographystyle{IEEEtran}

\end{document}